# Spatial-spectral FFPNet: Attention-Based Pyramid Network for Segmentation and Classification of Remote Sensing Images

Qingsong Xu

*Key laboratory of Mountain Hazards and Surface Process, Institute of Mountain Hazards and Environment, Chinese Academy of Sciences, Chengdu, 610041, China, and is with University of Chinese Academy of Sciences, Beijing 100049, China.*

Xin Yuan

*Bell Labs, Murray Hill, New Jersey 07974, USA.*

Chaojun Ouyang[*]

*Key laboratory of Mountain Hazards and Surface Process, Institute of Mountain Hazards and Environment, Chinese Academy of Sciences, Chengdu, 610041, China, and is with University of Chinese Academy of Sciences, Beijing 100049, China, and is with CAS Center for Excellence in Tibetan Plateau Earth Sciences, Chinese Academy of Sciences (CAS), Beijing 100101, China (e-mail: cjouyang@imde.ac.cn).*

Yue Zeng

*School of Economics and Management, Southwest Jiao Tong University, Chengdu 610031, China*

---

**Abstract**

We consider the problem of segmentation and classification of high-resolution and hyperspectral remote sensing images. Unlike conventional natural (RGB) images, the inherent large scale and complex structures of remote sensing images pose major challenges such as *spatial object distribution diversity* and *spectral information extraction* when existing models are directly applied for image classification. In this study, we develop an attention-based pyramid network for segmentation and classification of remote sensing datasets. Attention mechanisms are used to develop the following modules: *i*) a novel and robust attention-based *multi-scale fusion* method effectively fuses useful spatial or spectral information at different and same scales; *ii*) a *region pyramid attention* mechanism using region-based attention addresses the target geometric size diversity in large-scale remote sensing images; and *iii*) *cross-scale attention* in our adaptive atrous spatial pyramid pooling network adapts to varied contents in a feature-embedded space.

Different forms of feature fusion pyramid frameworks are established by combining these attention-based modules. First, a novel segmentation framework, called the heavy-weight spatial feature fusion pyramid network (FFPNet), is proposed to address the spatial problem of high-resolution remote sensing images. Second, an end-to-end spatial–spectral FFPNet is presented for classifying hyperspectral images. Experiments conducted on ISPRS Vaihingen and ISPRS Potsdam high-resolution datasets demonstrate the competitive segmentation accuracy achieved by the proposed heavy-weight spatial FFPNet. Furthermore, experiments on the Indian Pines and the University of Pavia hyperspectral datasets indicate that the proposed spatial–spectral FFPNet outperforms the current state-of-the-art methods in hyperspectral image classification.

**1. Introduction**

Supervised segmentation and classification are important processes in remote sensing image perception. Many socioeconomic and environmental applications, including urban and regional planning, hazard detection and avoidance, land use and land cover classification, and mapping and tracking, can be handled by using suitable remote sensing data and effective classifiers [1]. A great deal of data with different spectral and spatial resolutions is currently available for different applications with the development of modern remote sensing technology. Among these massive remote sensing data, high-resolution and hyperspectral images are two important types. Highresolution remote sensing images usually have rich spatial distribution information and a few spectral bands, which contain the detailed shape and appearance of objects [2]. Semantic segmentation is a powerful and promising scheme to assign pixels in high-resolution images with class labels [3, 11]. Hyperspectral images can capture hundreds of narrow spectral channels with an extremely fine spectral resolution, allowing accurate characterization of the electromagnetic spectrum of an object and facilitating a precise analysis of soils and materials [5]. Because each pixel can be considered a highdimensional vector and to be surrounded by local spatial neighborhood, supervised spatial–spectral classification methods are suitable for hyperspectral images.

However, segmentation or classification of different types of remote sensing images is an exceedingly difficult process, which includes major challenges of **spatial object distribution diversity** (Fig. 1) and **spectral information extraction**. Specifically, the following are the challenges with segmentation and classification of remote sensing images:

- Missing pixels or occlusion of objects: different from traditional (RGB) imaging methods, remote sensing examines an area from a significantly long distance and gathers information and images remotely. Due to the large areas contained in one sample and the effects of the atmosphere, clouds, and shadows, missing pixels or occlusion of objects are inevitable problems in remote sensing images.

- Geometric size diversity: the geometric sizes of different objects may vary greatly and some objects are small and crowded in remote sensing imagery because of the large area covered comprising different objects (e.g., cars, trees, buildings, roads in Fig. 1).

- High intra-class variance and low inter-class variance: this is a unique problem in remote sensing images and it inspires us to study superior methods aiming to effectively fuse multiscale features. For example, in Fig. 1, buildings commonly vary in shape, style, and scale; low vegetations and impervious surfaces are similar in appearance.

- Spectral information extraction: hyperspectral image datasets contain hundreds of spectral bands, and it is challenging to extract spectral information because of the similarity between the spectral bands of different classes and complexity of the spectral structure, leading to the Hughes phenomenon or curse of dimensionality [6]. More importantly, hyperspectral datasets usually contain a limited number of labeled samples, thus making it difficult to extract effective spectral information from hyperspectral images.

*1.1. Review of Semantic Segmentation of High-resolution Remote Sensing Images by Multiscale Feature Processing*

First, to solve the problem of spatial object distribution diversity in highresolution images, it is necessary to effectively extract and fuse features in multiple scales. Recently, deep-learning methods have shown excellent performance in remote sensing image processing, especially deep convolutional neural networks (DCNNs), which have strong ability to express multiscale features (such as a bottom-up feature pyramid obtained by multiple convolution operations [7]). To date, many models based on DCNNs for semantic segmentation of remote sensing images have been proposed. Sun and Wang [3] established a semantic segmentation scheme based on fully convolutional networks [8]. Wang et al. [9] proposed a gated network based on the information entropy of feature maps. This method can effectively integrate local details with contextual information. The cascaded convolutional neural networks [10, 4] were utilized for the segmentation of remote sensing images by successively aggregating contexts. Most recently, many multiscale context-augmented models [12, 13, 14] have been proposed to exploit contextual information in remote sensing images. Remote sensing target segmentation problems such as object occlusion, geometric size diversity, and small objects have attracted increasing research attention [15, 16, 17, 18].

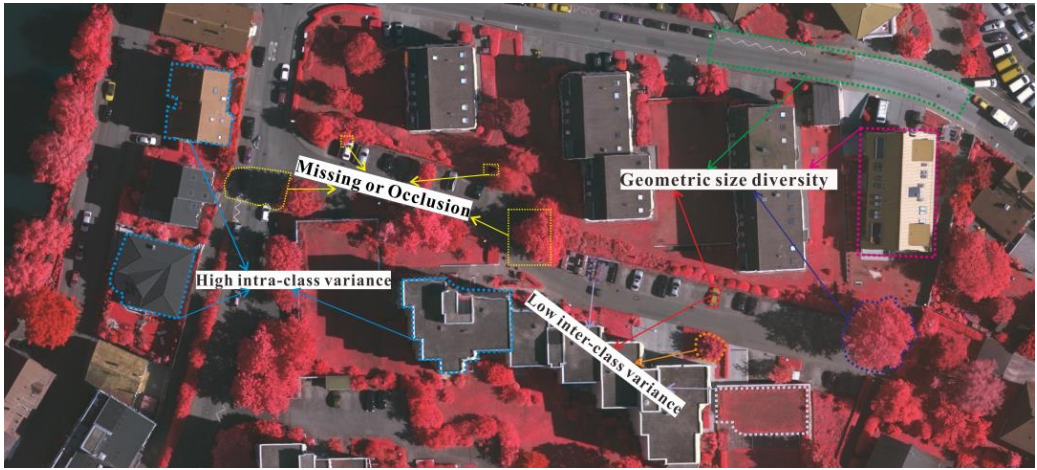

Figure 1: Challenges of object segmentation in spatial distribution in remote sensing images.

    Further analysis of these multiscale/contextual feature fusion models reveals that their common objective is to establish an effective feature attention weight fusion method. Attention mechanisms are widely used for various tasks such as machine translation [19], scene classification, and semantic segmentation. The non-local network [20] first adopts a self-attention mechanism as a submodule for computer vision tasks. Recently, many attentionreinforced mechanisms [21, 13, 22] have been proposed on the basis of nonlocal operation in semantic
4

segmentation. Attention U-Net [23] learns to suppress irrelevant areas in an input image while highlighting useful features for a specific task on the basis of cross-layer self-attention. CCNet [24] harvests the contextual information of all the positions in one image by stacking two serial criss-cross attention modules. ACFNet [25] is a coarse-to-fine segmentation network based on the attention class feature module, which can be embedded in any base network. Most recently, various self-attention mechanisms have proven to be effective for solving the problem of multiscale feature fusion in feature pyramid-based models [26, 27, 28, 29].

In summary, the above-mentioned multiscale feature fusion models based on attention mechanisms apply convolutional neural networks (CNNs) in three-band data, which have achieved significant breakthroughs in semantic segmentation. However, these models still cannot effectively solve the problem of spatial distribution diversity in remote sensing for the following reasons:

1) Most models only consider the fusion of two or three adjacent scales and do not further consider how to achieve the feature fusion of more or even all the different scale layers. Improved classification accuracy can be achieved by combining useful features at more scales.

2) Although a small part of the attention mechanism considers the fusion of more layers, it does not successfully solve the semantic gaps between high- and low-level features. The detailed analysis of different feature layers is discussed in Section 2.1.

3) The novel attention mechanisms based on self-attention mainly focus on spatial and channel relations for semantic segmentation (such as the non-local network [20]). Regional dependencies are not considered for the remote sensing images, and thus the relationship between object regions cannot be deepened.

*1.2. Review of Spatial–spectral Classification for Hyperspectral Images by Multiscale Feature Processing*

To solve the problem of spectral information extraction in hyperspectral images and enhance the classification performance, spatial–spectral classification methods have gained prominent application in hyperspectral image processing, mainly including handcrafted feature-based approaches [30, 31, 32, 33] and deep learning methods. Since deep learning methods (especially DCNNs) have proven to be more advantageous in feature extraction and representation compared with the traditional shallow learning method, this paper mainly focuses on deep spatial–spectral feature extraction and representation by multiscale feature processing in DCNNs. A review of DCNNbased classification methods for spatial–spectral approaches is given in [5], including 1D or 2D CNN [34, 35], 2D+1D CNN [36], and 3D CNN [37, 38, 39]. However, although these methods achieve promising performance for hyperspectral classification, they cannot fully extract and represent features, because they utilize the features of only the last convolutional layer for classification without considering multiscale features obtained by the previous convolutional layers. To this end, Zhao et al. [40] proposed a multiple convolutional layer fusion framework to fuse features extracted from different convolutional layers. The fusion process mainly involves the majority voting or direct concatenate mechanisms after applying the fully connected layer to each convolutional layer. The CNNs with multiscale convolution (MSCNNs) [41] are proposed to address the limited number of training samples and class differences in variance for hyperspectral images by extracting deep multiscale features. By conducting experiments on three popular hyperspectral images, Imani and Ghassemian [42] demonstrated that although feature fusion methods are time consuming, they can provide superior classification accuracy compared to other methods. Imani and Ghassemian [42] also showed that multiscale feature fusion is developed into one of the trends of hyperspectral image classification. Furthermore, attention mechanisms are used to extract and fuse contextual features. Haut et al. [43] is the first to develop a visual attention-driven mechanism for spatial–spectral hyperspectral image classification, which applies the attention mechanism to residual neural networks. Mei et al. [44] proposed a spatial–spectral attention network for hyperspectral image classification by the RNN and CNN both with the attention mechanism. However, these methods are only the initial application of multiscale fusion and the attention mechanism in hyperspectral datasets. There is still room for improvement in the following aspects in the area of hyperspectral image classification:

1) When dealing with hyperspectral spatial neighborhoods of the considered pixel, the semantic gap in multiscale convolutional layers is not considered, and simple fusion is not the most effective strategy.

2) The spectral redundancy problem is not considered sufficiently in the existing hyperspectral classification models. With regard to such a complex spectral distribution, there is exceedingly little work on extraction of spectral information from coarse to fine (multiscale) processing by different channel dimensions.

Bearing the above challenges in mind, in this study, we propose an **attention-based pyramid network** by using a self-attention mechanism flexibly. Our model utilizes attention mechanisms in the following three areas:

1) We propose attention-based *multiscale fusion* to fuse useful features at different and the same scales to achieve the effective extraction and fusion of spatial multiscale information and extraction of spectral information from coarse to fine scales.

2) We propose *cross-scale attention* in our adaptive atrous spatial pyramid pooling (adaptive-ASPP) network to adapt to varied contents in a feature-embedded space, leading to effective extraction of the context features.

3) A *region pyramid attention* module based on region-based attention is proposed to address the target geometric size diversity in large-scale remote sensing images.

Through different combinations of these attention modules, different forms of feature fusion pyramid frameworks (two-layer and three-layer pyramids) are established. First, a novel and practical segmentation model, called the heavy-weight spatial feature fusion pyramid network (FFPNet), is proposed to solve the spatial object distribution diversity problem in high-resolution remote sensing images. The heavy-weight spatial FFPNet is a three-level feature fusion pyramid built on the basis of region pyramid attention and attention-based multiscale fusion modules. Furthermore, boundary-aware (BA) loss [45] is used to train the heavy-weight spatial FFPNet in an end-toend manner. Second, a spatial–spectral FFPNet is developed to extract and integrate multiscale spatial features and multi-dimensional spectral features of hyperspectral images using the attention-based multiscale fusion module. The spatial–spectral FFPNet mainly consists of two modules: a light-weight spatial feature fusion pyramid (FFP) and a spectral FFP. The light-weight spatial FFP is a two-level pyramid, whose trainable parameters are less than one-third those of the heavy-weight spatial FFPNet. Thus, the light-weight module is suitable for a small number of labeled samples of the hyperspectral dataset. In addition, the spectral FFP, which is also a two-level pyramid, is proposed to better extract the spectral features from hyperspectral datasets by compressing spectral information from coarse to fine scales.

To evaluate the accuracy and efficiency of the proposed models, first, extensive experiments are conducted on two challenging high-resolution semantic segmentation benchmark datasets, namely the ISPRS Vaihingen dataset and the ISPRS Potsdam dataset. The local experimental results demonstrate that the heavy-weight spatial FFPNet outperforms other predominant DCNN-based models (DeepLabv3+ [46] considered as the baseline). In addition, the effectiveness and practicability of these novel attention-based mechanisms is demonstrated by conducting an ablation study. Furthermore, we apply the spatial–spectral FFPNet to two popular hyperspectral datasets, namely the Indian Pines dataset and the University of Pavia dataset. The experimental results (the well-known CNN model [38] considered as the baseline) indicate that the spatial–spectral FFPNet is more robust for a small number of training samples of the hyperspectral dataset and can obtain state-of-the-art results under different training samples. Our proposed spatial–spectral FFPNet has excellent ability to extract and express multiscale spatial and spectral information. It is worth noting that the spatial– spectral FFPNet with data enhancement is a better choice for hyperspectral image classification when the sample size is extremely small.

Overall, our proposed methods have set a new banner for the research on CNN models for high-resolution image segmentation and hyperspectral image classification. In addition, they may contribute to the development of attention mechanisms in CNNs.

## 2. Proposed Spatial–spectral FFPNet

*2.1. Overview*

In this study, we focus on the challenge of spatial and spectral distribution of remote sensing images in the "encoder–decoder" frameworks [48, 49, 50, 13, 9, 46]. The encoder part is based on a convolutional model to generate a feature pyramid with different spatial levels or spectral dimensions. Then, the decoder fuses multiscale contextual features. The interaction of adjacent scales can be formulated as

$$\mathbf{F}_l = \mathrm{H}(\mathbf{f}_l, \mathbf{f}_{l+1}), \qquad (1)$$

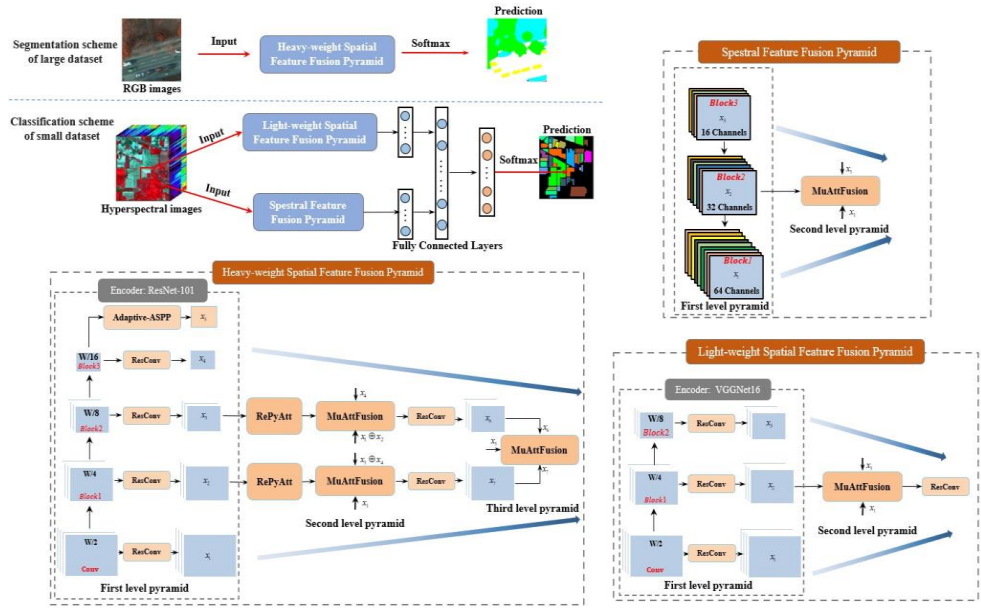

Figure 2: Proposed segmentation scheme of high-resolution images and classification scheme of hyperspectral images (upper left). For the heavy-weight spatial FFPNet, given a high-resolution image (3-band), ResNet-101 pretrained on ImageNet [47] is used as the backbone for feature extraction (middle, where $W$ denotes the height or width of the image). **First-level pyramid:** the features from four stages of the backbone are fed into ResConv to generate $x_1$, $x_2$, $x_3$, and $x_4$. The output of the backbone is fed into adaptive-ASPP by combining these context features by cross-scale attention to generate $x_5$. The intermediate features $x_2$ and $x_3$ are sent to RePyAtt based on region-based attention (middle). MuAttFusion is proposed to fuse the useful features at different scales and the same scales effectively. **Second-level pyramid:** $x_6$ and $x_7$ are generated from MuAttFusion (right), where $\oplus$ denotes the concatenation operation. **Third-level pyramid:** the final predicted segmentation map is generated after using MuAttFusion for $x_6$, $x_7$, and $x_5$ again. The detailed configurations of the heavy-weight spatial FFPNet are described in Section 2.5. Furthermore, BA loss is used to train the FFPNet in an end-to-end manner. For the spatial–spectral FFPNet, given a hyperspectral image with the size of $p \times H \times W$, where $p$ is the number of spectral bands, the image is sent to the light-weight spatial FFP and the spectral FFP modules simultaneously. First, the light-weight spatial FFP is a two-level pyramid, and VGGNet16 pretrained on ImageNet [47] is used as the backbone. Notably, the initial parameters of the first convolutional layer in the pretrained network are copied until the $p$-channel inputs are attained. **First-level pyramid:** the features from three blocks of the backbone are fed into ResConv to generate $x_1$, $x_2$, and $x_3$; **Second-level pyramid:** MuAttFusion is used to fuse the useful features from $x_1$, $x_2$, and $x_3$ to generate the multi-scale spatial fusion feature. Second, the spectral FFP module is also a two-level pyramid, which can be divided into three stages by different channels with depths of 64, 32, and 16, respectively. Every stage contains $3 \times 3$ and $1 \times 1$ convolutional layers. MuAttFusion is then harnessed to extract and combine these features of different stages. Finally, fully connected layers are used to effectively merge multiscale spatial and spectral features

and predict the class of all pixels. The detailed description of the spatial–spectral FFPNet is presented in Section 2.6.

where $\mathbf{F}_l$ is the fused feature at the $l^{th}$ level, H represents a combination of multiplication [49, 51], weight sum [52], concatenation [50], attention mechanism [48, 53, 26], and other operations [9].

However, these operations cannot solve the problem of multiscale feature fusion of objects in remote sensing images. The main reason is that the feature maps from the lower layers are of high resolution and may have excessive noise, resulting in insufficient spatial details for high-level features. Further, these integrated operations may suppress necessary details in the low-level features, and most of these fusion methods do not consider the large semantic gaps between the feature pyramids generated by the encoder. Furthermore, these operations do not consider effective extraction and fusion of multiscale spectral information in hyperspectral images.

Therefore, we propose a novel multi-feature fusion model based on attention mechanisms in this paper. Current attention mechanisms [54, 13, 21, 23] are based on the non-local operation [20], which usually deal with spatial pixel and channel selections. These mechanisms cannot achieve regional dependencies of objects and cannot effectively extract and integrate multiscale features in remote sensing images. To address these issues, three novel attention modules are proposed: 1) A region pyramid attention (RePyAtt) module is proposed to effectively establish dependencies between different region features of objects and relationships between local region features by using a self-attention mechanism on different feature pyramid regions; 2) An adaptive-ASPP module aims to adaptively select different spatial receptive fields to tackle large appearance variations in remote sensing images by adding an adaptive attention mechanism to the ASSP [55, 46]; 3) A multiscale attention fusion (MuAttFusion) module is proposed to fuse the useful features at different scales and the same scales effectively.

As shown in Fig. 2, segmentation and classification schemes of remote sensing images are achieved through the different combinations of the proposed attention modules. First, for high-resolution images, most of the information is concentrated in spatial dimensions. The proposed heavy-weight spatial FFPNet segmentation model solves the spatial object distribution diversity problem in remote sensing images. We adopt ResNet-101 [56] pretrained on ImageNet [47] as the backbone of the segmentation model. A three-level feature fusion pyramid is designed as shown in Fig. 2. In addition, the residual convolution (ResConv) module is used as the basic processing unit, while the adaptive-ASPP module is used to adaptively combine the context features generated from the ResNet-101 and ResConv.

Second, for hyperspectral images, the proposed spatial–spectral FFPNet extracts and integrates multiscale spatial and spectral features. Recalling Fig. 2, the spatial–spectral FFPNet includes three parts: 1) multiscale spatial feature extraction with the light-weight spatial FFP; 2) multi-spectral feature extraction with the spectral FFP; 3) fusion of spatial and spectral features as well as classification prediction with fully connected layers. Specifically, the light-weight spatial FFP module is a shallow classification framework, which uses the blocks of VGGNet-16 [57]. It only has a two-level feature fusion pyramid based on MuAttFusion. In comparison, the trainable parameters of the light-weight spatial FFP module are less than one-third those of the heavy-weight spatial FFPNet. This is because of the small number of labeled samples in hyperspectral datasets. The more parameters a model has, the greater its capacity, but also more labeled data needed to prevent overfitting. Similarly, the spectral FFP module has a two-level feature fusion pyramid based on MuAttFusion, which reduces the amount of parameters while capturing as much spectral information as possible.

*2.2. Region Pyramid Attention Module*

Currently, the soft attention-based methods mainly aim to capture longrange contextual dependencies on the basis of the non-local mechanism and its variants. However, the geometric size of different objects in remote sensing images varies significantly, so it is challenging to achieve regional dependencies of objects using existing models. Inspired by the ideas of the feature pyramid, we propose the *region pyramid* to address the target scale diversity. After this, we combine the region pyramid and self-attention to effectively establish dependencies between different object region features and relationships between local region features. We illustrate our approach via a simple schematic in Fig. 3.

*2.2.1. Region pyramid*

We partition the input feature maps into different regions via a chunk operation. The region block size defined in this article are {single pixel level, 8 × 8 level, 4 × 4 level, 2 × 2 level, and 1 × 1 level}. In addition, we conduct an ablation study on different combinations as detailed in Section 3.3.2. For each group of the pyramid, we first feed the region blocks into a global pooling layer to obtain the regional representations. Then, we concatenate the representations of the region block to generate a regional representation of the whole input feature. It is worth noting that the single pixel level is directly sent to the self-attention module without the global pooling operation.

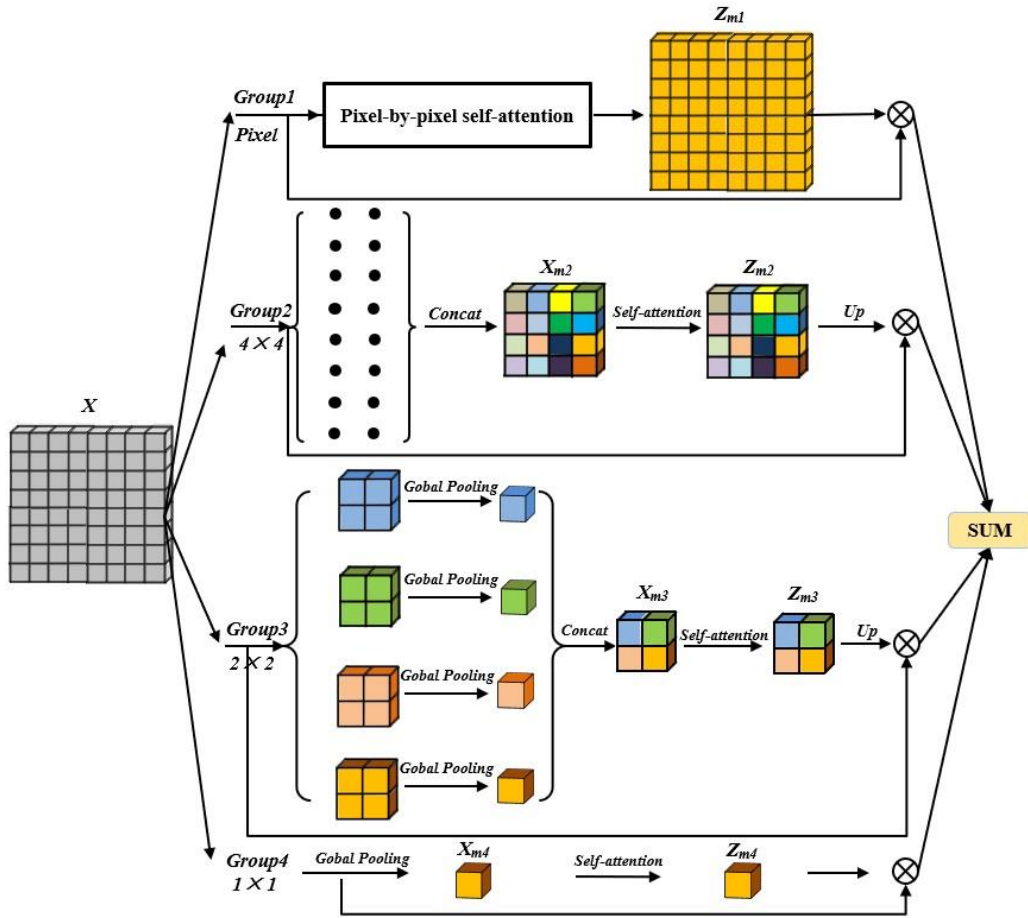

Figure 3: Proposed RePyAtt Module. We first generate a region pyramid by partitioning the input feature maps (left) into four groups and employing the selfattention mechanism to extract the regional dependence. Finally, the output of the RePyAtt module is obtained by summation ('SUM') of different region groups.

*2.2.2. Self-attention on the regional representation*

To exploit more explicit regional dependencies of objects, we compute the self-attention representations within the regional representation. Selfattention consists of one 3 × 3 convolution and one 1 × 1 convolution, with the number of channels *F/2* and 1, respectively, where *F* denotes the number of channels of the input feature maps. Further experiments show that the parallel form of attention-weighted representations of different region groups can effectively enhance the dependencies across different region features and the relationships between local region features better than pixel-wise and channel-wise self-attention operators.

As illustrated in **Group 3** of Fig. 3, we first divide the input feature **X** into $G$ ($2 \times 2$) partitions. Then, we concatenate the point statistics after global pooling to obtain the regional representation $\mathbf{X}_{m3} \in \mathbf{R}^{F \times G}$. We apply self-attention on $\mathbf{X}_{m3}$ as follows:

$$A_{m3} = \text{softmax}(W_1 * \mathbf{X}_{m3}), \qquad Z_{m3} = A_{m3} f(\mathbf{X}_{m3}) + \mathbf{X}_{m3}, \qquad (2)$$

where $A_{m3} \in \mathrm{R}^{1 \times G}$ is an attention matrix based on the global information across all spectral bands, and $Z_{m3} \in \mathrm{R}^{F \times G}$ is the weighted output features. $W_1$ denotes the combination operation of one $3 \times 3$ convolution and $1 \times 1$ convolution. $f(.)$ represents $1 \times 1$ convolution and $*$ denotes convolution.

Finally, the output of the RePyAtt module is obtained by the weighted sum of different region groups, which is formulated as

$$\sum_{i=1}^{M} Up(Z_{mi}) \otimes \mathbf{X}, \qquad (3)$$

where $M$ represents the total number of groups in the region pyramid, $\otimes$ denotes region-wise multiplication, and $Up(.)$ is the upsampling layer using nearest interpolation.

*2.3. Multi-scale Attention Fusion*

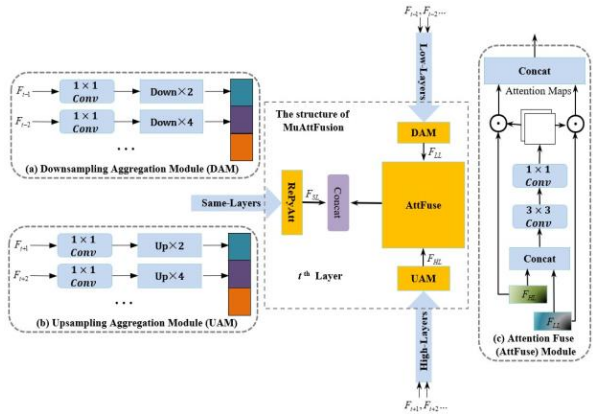

Figure 4: Proposed MuAttFusion module. It selectively fuses the same-layer, higher-layer, and lower-layer features using an adaptive attention method.

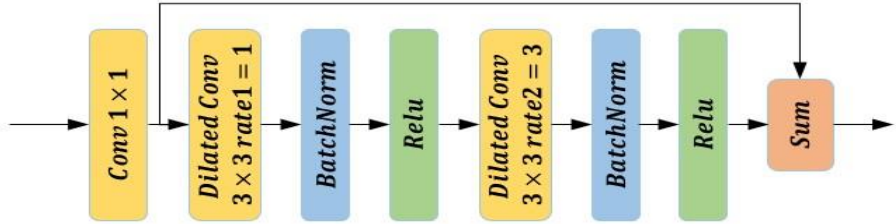

Figure 5: ResConv module used to refine the features and reduce network parameters.

The main task of the proposed MuAttFusion module is to effectively integrate multiscale spatial and spectral features of different objects in remote sensing images. MuAttFusion selectively fuses same-layer, high-layer, and low-layer features by an adaptive attention method as shown in Fig. 4.

### 2.3.1. Higher- and lower-scales

The lower-layer branch propagates spatial information from the lower layers ($T < t$) to the current layer ($t$) by the downsampling aggregation module (DAM). As shown in Fig. 4(a), to minimize memory consumption, we first use a 1 × 1 convolutional layer to compress the incoming feature maps. To achieve a consistent size for all feature maps, low-level features are downsampled to the feature size of the current layer by using bilinear interpolation. To fully use the entire feature information, all lower-layer feature maps are concatenated. Introducing the lower layers into the current layer inadvertently passes noise as well. To tackle this, high-level ($T > t$) contextual information is simultaneously propagated into the current layer by the upsampling aggregation module (UAM). The UAM structure is similar to that of the DAM, as shown in Fig. 4(b).

### 2.3.2. Attention fuse module

The lower-layer features, although refined, may contain some unnecessary background clutter, whereas in the higher-layer features, the detailed information may be oversuppressed in the current layer. To address these issues, we introduce an attention fuse (AttFuse) module, shown in Fig. 4(c). This module combines features of these two branches by adaptive attention weights. Consider the two feature maps $F_{LL}$ and $F_{HL}$; the attention module concatenates them and feeds them through a set of convolution layers (3 × 3 conv and 1 × 1 conv) and a sigmoid layer to produce an attention map with two channels, with each channel specifying the importance of the corresponding feature map. The attention maps are calculated as follows: $A_f$ = *sigmoid* (*Concat*[$F_{LL},F_{HL}$]) . The attention maps thus generated are then multiplied element-wise to produce the final higher- and lower-layer fusion feature maps: $F_f = A_f^1 \odot F_{LL} + A_f^2 \odot F_{HL}$. $F_f$ is a powerful and enriched multiscale feature by combining the advantages of lower-layer features $F_{LL}$ and $F_{HL}$.

Finally, the output feature $\tilde{F}_t$ of the MuAttFusion module is then fused with the same-layer features by the RePyAtt module: $\tilde{F}_t$ = *Concat*[$F_{SL},F_f$]. It is worth noting that for the light-weight model, the output feature and the same-layer features are directly fused to reduce the model parameters.

To further refine the features and reduce network parameters. ResConv shown in Fig. 5 is introduced. The ResConv block consists of one 1 × 1 convolution and two 3 × 3 *dilated* convolution, with rates = 1 and 3. The 1×1 convolution reduces the network channel, thereby reducing the network parameters. Two 3×3 dilated convolution can deepen the network to enhance its ability to capture sophisticated features.

*2.4. Adaptive-ASPP Module*

Objects within a remote sensing image typically have different sizes. Existing multiple branch structures such as ASPP [55, 46] and DenseASPP [58] are developed to learn features using filters of different dilation rates in order to adapt to the scales. However, these approaches ignore the same problem : different local regions may correspond to objects with different scales and geometric variations. Thus, **spatial adaptive filters** are desired for different scales to tackle large feature variations in remote sensing images.

Toward this end, inspired by the MuAttFusion module described in Section 2.3, an adaptive-ASPP is designed to adapt to varied contents. The core of adaptive-ASPP is to adjust the combination weights for different contents in a feature-embedded space. The CASINet [29] was proposed to solve this problem; it first uses a non-local operation to achieve the adaptive information interaction across scales. However, the non-local operation [20] was used to exploit the long-range contexts for feature refinement and its calculation cost is high. The non-local operation is not applicable for cross-scale attention problems; this is also verified by our experiments. Different from the non-local operation, we propose a novel cross-scale attention (CrsAtt) module based on the self-attention mechanism.

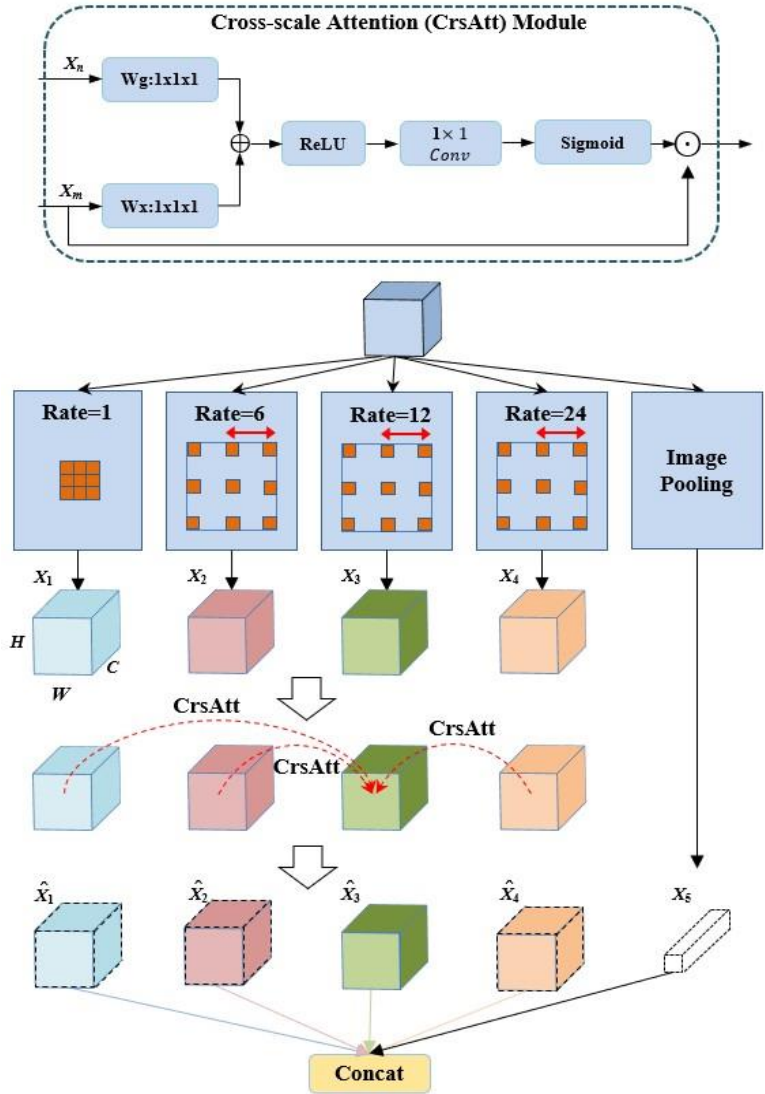

Figure 6: Structure of the proposed adaptive-ASPP module. It is designed to adjust the combination weights for varied contents in a feature-embedded space by using the proposed cross-scale attention (CrsAtt) module (top).

*2.4.1. Cross-scale attention module*

The structure of the proposed CrsAtt module is shown in the top of Fig. 6. CrsAtt first uses two different scales to obtain the attention coefficients; then, it adaptively adjusts different scale feature weights by element-wise multiplication of the input scale feature maps and attention coefficients.

As depicted in Fig. 6, consider five intermediate feature maps, $\{X_1, X_2, X_3, X_4, and X_5\}$, obtained from five branches of the ASPP with each $X_i \in R^{H \times W \times C}$ (except $X_5$, which is obtained by image pooling of features). Information interaction is performed across each scale feature of four scales $\{X_1, X_2, X_3, X_4\}$, with each scale being a feature node. Then, CrsAtt operations are performed on the four features. The feature of the $i^{th}$ scale is calculated as

$$A_{ji} = \sigma_2 \left( \varphi^T \left( \sigma_1 \left( W_g^T * X_j + W_x^T * X_i \right) \right) \right)$$
$$\hat{X}_i = \sum_{\substack{j=1 \\ j \neq i}}^{j=4} A_{ji} X_i + X_i, \quad (4)$$

where $\sigma_1 = \max(0, x)$ and $\sigma_2 = \frac{1}{1+e^{-x}}$ correspond to ReLU and Sigmoid activation functions, respectively. $*$ denotes channel-wise $1 \times 1 \times 1$ convolutional layer parameterized by $W_x \in R^{C \times C_{int}}$, $W_g \in R^{C \times C_{int}}$. In addition, $\phi^T \in R^{C_{int} \times 1}$ is computed using channel-wise $1 \times 1 \times 1$ convolutions.

Finally, the output of the adaptive-ASPP is obtained by concatenating $\{\hat{X}_1, \hat{X}_2, \hat{X}_3, \hat{X}_4,$ and $X_5\}$.

*2.5. Heavy-weight Spatial FFPNet Model*

The heavy-weight spatial FFPNet model is achieved through the combination of the three attention-based modules introduced in the previous sections. The configurations of the three-level heavy-weight FFPNet is shown in Table 1. Concretely, consider an input image $\mathbf{X} \in R^{C \times H \times W}$, in which $C$, $H$, and $W$ denote the number of channels, height, and width of the image, respectively. First, the image is fed it into the ResNet-101 [56] pretrained on the ImageNet dataset [47] to generate different scale feature maps. In the first level of the pyramid, the features from the four stages of the backbone are fed into ResConv to generate different scale feature maps $x_1$, $x_2$, $x_3$, and $x_4$, with 256 channels, respectively. In addition, the output of the backbone is fed into the adaptive-ASPP module to generate the feature map $x_5$ to adaptively combine these context features. In the second level of the pyramid, the intermediate features $x_2$ and $x_3$ are sent to RePyAtt based on region-based attention; $x_6$ and $x_7$ are then generated after MuAttFusion. In the third level of the pyramid, the final predicted segmentation map is generated after using MuAttFusion for $x_6$, $x_7$, and $x_5$ again. Furthermore, BA loss [45] is utilized to train the heavy-weight spatial FFPNet in an end-toend manner to optimize the model parameters. By the simple modification of cross entropy loss, the BA loss is utilized to solve the issue that the pixels surrounding the boundary are hard to predict.

Table 1: Three-level heavy-weight spatial FFPNet configurations. The module parameters are denoted as " module name(receptive fields of different convolutions)-number of modules-number of module output channels". Note that some complex modules only give the module name

| Level | Detailed configurations |
|---|---|
| First-level pyramid | $x_1$: Conv2d(7×7)-1-64 + ResConv(1×1+3×3+3×3)-1-256 |
| | $x_2$: Maxpool + Block1(1×1+3×3+1×1)-3-256 + ResConv(1×1+3×3+3×3)-1-256 |
| | $x_3$: Block2(1×1+3×3+1×1)-4-512 + ResConv(1×1+3×3+3×3)-1-256 |
| | $x_4$: Block3(1×1+3×3+1×1)-23-1024 + Block4(1×1+3×3+1×1)-3-2048 + ResConv(1×1+3×3+3×3)-1-256 |
| | $x_5$: Adaptive-ASPP |
| Second-level pyramid | $x_7$: RePyAtt + MuAttFusion($x_1,x_2,x_3,x_4$) + ResConv(1×1+3×3+3×3)-1-256 |
| | $x_6$: RePyAtt + MuAttFusion($x_1,x_2,x_3,x_4$) + ResConv(1×1+3×3+3×3)-1-256 |
| Third-level pyramid | MuAttFusion($x_5,x_6,x_7$) |
| Parameter: | 78.8 million |

Table 2: Two-level light-weight spatial feature fusion pyramid configurations. The module parameters are denoted as "module name(receptive fields)-number of modules-number of module output channels".

| Level | Detailed configurations |
|---|---|
| First-level pyramid | $x_1$: Conv2d(3×3)-2-64 + Conv2d(3×3)-2-128 + Maxpool + ResConv(1×1+3×3+3×3)-1-256 |
| | $x_2$: Conv2d(3×3)-3-256 + Conv2d(3×3)-3-512 + Maxpool + ResConv(1×1+3×3+3×3)-1-256 |
| | $x_3$: Conv2d(3×3)-3-512 + Maxpool + ResConv(1×1+3×3+3×3)-1-256 |
| Second-level pyramid | MuAttFusion($x_1,x_2,x_3$) + ResConv(1×1+3×3+3×3)-1-256 |
| Parameter | 24.8 million |

*2.6. Spatial–spectral FFPNet Model*

To maximize the use of hyperspectral spatial and spectral information, instead of dealing with the hypercube as a whole, the proposed spatial–spectral FFPNet model includes two CNN modules: the light-weight spatial FFP module for learning multiscale spatial features and the spectral FFP module for extracting spectral features along multiple dimensions. The features from the two modules are then concatenated and fed to a fully connected classifier to perform spatial–spectral classification.

*2.6.1. Light-weight spatial feature fusion pyramid module*

The light-weight spatial FFP module is a relatively shallow spatial feature extraction framework, which only uses VGGNet-16 [57] as the backbone; the configurations of two-level light-weight spatial FFP module is shown in Table 2. Compared with the heavy-weight spatial FFPNet, the light-weight one only has 24.8 million trainable parameters owing to the small number of labeled hyperspectral samples. Furthermore, MuAttFusion is utilized to fuse the useful features from $x_1$, $x_2$, and $x_3$, generated by the backbone after the execution of ResConv.

*2.6.2. Spectral feature fusion pyramid module*

As shown in Fig. 2, the spectral module can use multiple convolutional kernels to automatically extract features from fine to coarse scales as convolutional layer progresses. Similarly, to solve the problem of spectral redundancy of hyperspectral images, the spectral information can be compressed by different channel dimensions from coarse to fine scales, and the useful features in the multiple scales are selected and merged by the attention mechanism. Thus, the spectral FFP module extracts the spectrum features of hyperspectral data more effectively. The configurations of the two-level spectral FFP module are presented in Table 3. Specifically, the multiscale features can be divided into three stages by different channels, with depths of 64, 32, and 16. Every stage contains a 3×3 convolutional layer to reduce the dimension of features and a 1 × 1 convolutional layer to further enhance the expression ability of spectral features. MuAttFusion is then harnessed to extract and combine useful features generated from the three stages.

Table 3: Two-level spectral feature fusion pyramid configurations. The module parameters are denoted as "module name(receptive fields)-number of modules-number of module output channels".

| Level | Detailed configurations |
|---|---|
| First-level pyramid | $x_1$: Conv2d(3×3+1×1)-1-64 |
| | $x_2$: Conv2d(3×3+1×1)-1-32 |
| | $x_3$: Conv2d(3×3+1×1)-1-16 |
| Second-level pyramid | MuAttFusion($x_1,x_2,x_3$) |
| Parameter | 0.20 million |

*2.6.3. Merge*

In the spatial–spectral FFPNet, the last step is the combination of the output features of the light-weight spatial FFP and spectral FFP modules. The overall framework is shown in Fig. 2. To effectively merge the spatial and spectral features as well as express the fused spatial–spectral features, first, the multiscale spatial features generated by the light-weight spatial FFP module and the multi-dimensional spectral features extracted by the spectral FFP module are converted into a one-dimensional tensor by a fully connected layer with the ReLU activation function. Then, the two types of features are directly merged through concatenation. Finally, another fully connected layer with the ReLU activation function is used to further refine and represent the combined spectral–spatial features, and a softmax activation layer predicts the probability distribution of each class. It is worth noting that owing to the fully connected layers, the spatial–spectral FFPNet has excellent robustness in terms of the inference to slide over the entire image.

Furthermore, to prevent the model from overfitting in case of limited hyperspectral datasets, the dropout method is used for the fully connected layers. Specifically, the dropout method randomly selects hidden neurons as zero with a probability of 0.5. These dropped neurons will not play a role in the forward and backward processes of the model.

## 3. Experiments

Numerical experiments were carried out on two high-resolution remote sensing datasets, namely ISPRS Vaihingen[1] dataset and ISPRS Potsdam[2] dataset, to validate the effectiveness of the heavy-weight spatial FFPNet segmentation model. Furthermore, in order to evaluate the performance of our newly presented spatial–spectral FFPNet classification architecture, two popular hyperspectral image datasets, namely the AVIRIS Indian Pines dataset[3] and the University of Pavia dataset[4] were utilized.

### 3.1. Dataset Description and Baselines
### 3.1.1. High-resolution datasets

The Vaihingen dataset consists of 3-band IRRG[5] image data acquired by airborne sensors. There are 33 images with a spatial resolution of 9 cm. The average size of each image is 2494×2064 pixels. All datasets are labeled into the five foreground classes (impervious surfaces, buildings, low vegetation, trees, and cars) and one background class (see Fig. 7). Following the setup in the online test, 16 images were used as a training set, while the remaining 17 images[6] were used as a model testing set. We randomly sampled the 512 × 512 patches from the 33 images and the images were processed at the training stage with normalization, random horizontal flip, and Gaussian blur. Finally, 6200 images were generated in the training set and 1000 images in the testing set. The Potsdam dataset is composed of 38 images with a spatial resolution of 5 cm and consists of 3-band IRRG image data. The size of all images is 6000×6000 pixels, which are annotated with pixel-level labels of six classes corresponding to the Vaihingen dataset. To train and evaluate the heavy-weight spatial FFPNet, we also followed the data partition method used in benchmark methods. 24 images were selected as a training set and 14 images[7] as a testing set. We randomly sampled the 512 × 512 patches from the original images and generated 14000 patches for the training set and 3000 patches for the testing set. Similar to the Vaihingen dataset, the patches were processed at the training stage with normalization, random horizontal flip, and Gaussian blur.

---

[1] http://www2.isprs.org/commissions/comm3/wg4/2d-sem-label-vaihingen.html
[2] http://www2.isprs.org/commissions/comm3/wg4/2d-sem-label-potsdam.html
[3] http://www.ehu.eus/ccwintco/index.php/Hyperspectral_Remote_Sensing_Scenes#Indian_Pines
[4] http://www.ehu.eus/ccwintco/index.php/Hyperspectral_Remote_Sensing_Scenes#Pavia_University_scene
[5] Infrared, Red and Green

*3.1.2. Hyperspectral datasets*

The AVIRIS Indian Pines (IP) dataset is gathered by the AVIRIS sensor. The image contains 224 spectral channels in the 400–2500 nm region of the visual and infrared spectra. As a conventional setup, 24 spectral bands were removed owing to noise and the remaining 200 bands were utilized for the experiments. The image is of size 145×145 with a spatial resolution of 20 m per pixel, and its ground truth contains sixteen different land-cover classes, which is shown in Fig. 7. 10249 pixels were selected for manual labeling according to the ground truth map. The University of Pavia (UP) dataset is recorded by the ROSIS-03 sensor. The image consists of 610 × 340 pixels and 115 bands with a spectral coverage ranging from 0.43 to 0.86 μm. After removing noisy bands, 103 bands were used. Nine classes of land covers were considered in the ground truth of this image, which are shown in Fig. 7.

*3.1.3. Baselines*

In order to evaluate the heavy-weight spatial FFPNet segmentation model, we chose DeepLabv3+ [46] as our baseline. DeepLabv3+ fuses multiscale features by introducing low-level features to refine high-level features; thus, state-of-the-art

---

[6] Image IDs: 2, 4, 6, 8, 10, 12, 14, 16, 20, 22, 24, 27, 29, 31, 33, 35, 38

[7] Image IDs: 02 13, 02 14, 03 13, 03 14, 04 13, 04 14, 04 15, 05 13, 05 14, 05 15, 06 13, 06 14, 06 15, 07 13

performance is achieved on many public datasets. Furthermore, for the spatial–spectral classification model, a generally recognized deep convolutional neural network proposed by Paoletti et al. [38] was utilized as the baseline for hyperspectral image classification. The CNN is a 3-D network using spatial and spectral information, which performs on hyperspectral datasets accurately and efficiently.

### 3.2. Evaluation Metrics

To evaluate the performance of the proposed models for segmentation and classification of remote sensing images, F1 score, Overall Accuracy (OA), Average Accuracy (AA), mean Intersection over Union (mIoU), and Kappa coefficient were used. First, for the measurement of heavy-weight spatial FFPNet performance, the F1 score was calculated for the foreground object classes and for a comprehensive comparison of the heavy-weight spatial FFPNet with different models, OA for all categories including background and mIoU were adopted. In addition, following a previous study [27], the ground truth with eroded boundaries was utilized for the evaluation in order to reduce the impact of uncertain border definitions. Second, AA, OA, and Kappa coefficient were used to measure the performance of spatial–spectral FFPNet classification. More importantly, the average results of three runs of all experiments of training and testing sets were calculated.

### 3.3. Heavy-weight Spatial FFPNet Evaluation on High-resolution Datasets
### 3.3.1. Implementation details

The stochastic gradient descent (SGD) was employed as the optimizer of the heavy-weight spatial FFPNet with momentum = 0.9 and weight decay = $5e - 4$. The initial learning rate = $2.5e - 4$, a poly learning rate policy was used for the optimizer. The mini-batch size was set to 4 and the maximum epoch was 10. In addition, the batch normalization and the ReLU

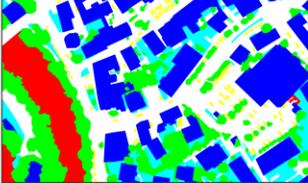

| Color | Land-cover type | | Color | Land-cover type | Samples | | Color | Land-cover type | Samples |
|---|---|---|---|---|---|---|---|---|---|
| 🟥 | Background | | | Background | 10776 | | | Background | 164624 |
| | Impervious surfaces | | | Alfalfa | 46 | | | Asphalt | 6631 |
| | | | | Corn-notill | 1428 | | | | |
| 🟦 | Buildings | | | Corn-mintill | 830 | | | Meadows | 18649 |
| | | | | Corn | 237 | | | | |
| 🟦 | Low vegetation | | | Grass-pasture | 483 | | | Gravel | 2099 |
| | | | | Grass-trees | 730 | | | | |
| 🟩 | Trees | | | Grass-pasture-mowed | 28 | | | Trees | 3064 |
| | | | | Hay-windrowed | 478 | | | | |
| 🟨 | Cars | | | Oats | 20 | | | Painted metal sheets | 1345 |
| | | | | Soybeans-notill | 972 | | | Bare Soil | 5029 |
| | | | | Soybeans-mintill | 2455 | | | | |
| | | | | Soybeans-clean | 693 | | | Bitumen | 1330 |
| | | | | Wheat | 205 | | | | |
| | | | | Woods | 1265 | | | Self-Blocking Bricks | 3682 |
| | | | | Bldg-grass-tree-drives | 386 | | | | |
| | | | | Stone-steel towers | 93 | | | Shadows | 947 |
| **Total image samples (Vaihingen/Postdam)** | 7200 / 17000 | | **Total pixel samples** | | 21025 | | **Total pixel samples** | | 207400 |

Figure 7: Ground-truth images of different datasets and the number of image samples for high-resolution datasets (Vaihingen and Potsdam) and pixel samples for hyperspectral datasets (the IP and UP datasets)

function were used in all layers, except for the output layers, where softmax units were used. Furthermore, inspired by the baseline model (Deeplabv3+), the dropout method (with probability = 0.5 and 0.1) was employed in the last layer of the decoder module to effectively avoid overfitting. We used Pytorch for implementation on a high-performance computing cluster, with one NVIDIA Titan RTX 24GB GPU. The average training time for each experiment was approximately 20 h.

*3.3.2. Experiments on Vaihingen dataset*

***Ablation study.***
In the proposed heavy-weight spatial FFPNet, three novel attention based modules extract and integrate multiscale features adaptively and effectively in remote sensing images. To verify the effectiveness of these attention-based modules, extensive experiments in different settings were conducted and the results are listed in Table 4. In addition, to study the adaptability of different combinations of region sizes to object features, we investigated different combinations of groups in the RePyAtt module, and the results are presented in Table 5.

As can be seen in Table 4, the three novel attention modules result in significant improvement compared to the baseline (Deeplabv3+ with ResNet101). Specifically, the use of the feature fusion pyramid framework yields an OA of 90.64% and an mIoU of 80.37%, which are 2.55% and 7.54% improvement, respectively, over the values yielded by the baseline. However, employing the ASPP module in the framework can lead to a slight decline on the model performance. This result is mainly because the ASPP module cannot solve well the issue of context feature fusion for geometric variations in remote sensing images. By contrast, the adaptive-ASPP adapts to varied contents by the CrsAtt module, thus improving the performance over the baseline by 2.82% and 8.51% in terms of OA and mIoU, respectively. Therefore, it is demonstrated that the adaptive-ASPP can be widely used in other related models in case of large appearance variations. Furthermore, the introduction of the BA loss can improve the performance by approximately 0.20% and 0.51% in terms of OA and mIoU compared with when the CE loss is used. Overall, the novel heavy-weight spatial FFPNet has great benefit in dealing with the spatial object distribution diversity challenge in remote sensing images.

We further studied the effects of different combinations of groups in the RePyAtt module. Table 5 shows that the performance is optimal and robust when the combination is set to { 4 × 4 level, 2 × 2 level, and 1 × 1 level}, in which the best OA and mIoU of 90.91% and 81.33%, respectively, are achieved. In addition, it can be observed that more combinations of groups do not necessarily result in better model performance. Thus, an optimal combination can be used to more effectively achieve region-wise dependencies of objects, resulting in improved model performance.

Table 4: Results of the ablation study on the Vaihingen testing dataset; the values in **bold** are the best. All results are the average of three runs with maximum epoch = 10 and mini-batch size = 4.

| Method | RePyAtt | MuAttFusion | Adaptive-ASPP | ASPP | CE Loss | BA Loss | OA(%) | mIoU(%) |
|---|---|---|---|---|---|---|---|---|
| ResNet-101 Baseline (Deeplabv3+) | | | | | √ | √ | 88.09 | 72.83 |
| ResNet-101+RePyAtt+MuAttFusion+BA | √ | √ | | | | √ | 90.64 | 80.37 |
| ResNet-101+RePyAtt+MuAttFusion+ASPP+BA | √ | √ | | √ | | √ | 90.37 | 79.96 |
| ResNet-101+RePyAtt+MuAttFusion+Adaptive-ASPP+BA | √ | √ | √ | | | √ | **90.91** | **81.33** |
| ResNet-101+RePyAtt+MuAttFusion+Adaptive-ASPP+CE | √ | √ | √ | | √ | | 90.71 | 80.82 |

Table 5: Results of the ablation study with different combinations of groups in the RePyAtt module; the values in **bold** are the best. All results are the average of three runs with maximum epoch = 10 and mini-batch size = 4.

| Pyramid combinations | OA( % ) | mIoU(%) |
|---|---|---|
| {single pixel,8,4,2,1} | 90.28 | 79.63 |
| {single pixel,4,2,1} | **90.91** | **81.33** |
| {single pixel,2,1} | 90.49 | 80.08 |
| {single pixel,1} | 90.66 | 80.39 |

Table 6: Experimental results on the Vaihingen dataset; the values in **bold** are the best.

| Method | Imp. Surf. | Build. | Low veg. | Tree | Car | mean F1(%) | OA(%) | mIoU(%) |
|---|---|---|---|---|---|---|---|---|
| FCNs [8] | 88.11 | 91.36 | 77.10 | 85.70 | 75.03 | 83.46 | 85.73 | 72.12 |
| DeepLabv3 [55] | 87.75 | 92.04 | 77.47 | 85.85 | 65.21 | 81.66 | 86.48 | 70.05 |
| UZ 1 [59] | 89.20 | 92.50 | 81.60 | 86.90 | 57.30 | 81.50 | 87.30 | - |
| DeepLabv3+ [46] | 90.03 | 93.13 | 79.08 | 87.09 | 68.94 | 83.65 | 88.09 | 72.83 |
| S-RA-FCN [13] | 91.47 | 94.97 | 80.63 | 88.57 | 87.05 | 88.54 | 89.23 | - |
| ONE 7 [60] | 91.00 | 94.50 | 84.40 | 89.90 | 77.80 | 87.52 | 89.80 | - |
| DANet [21] | 91.63 | 95.02 | 83.25 | 88.87 | 87.16 | 89.19 | 89.85 | 80.53 |
| GSN5 [9] | 91.80 | 95.00 | 83.70 | 89.70 | 81.90 | 88.42 | 90.10 | - |
| DLR 10 [61] | 92.30 | 95.20 | 84.10 | **90.00** | 79.30 | 88.18 | 90.30 | - |
| PSPNet [62] | 92.79 | **95.46** | 84.51 | 89.94 | **88.61** | **90.26** | 90.85 | **82.58** |
| Heavy-weight Spatial FFPENet | **92.80** | 95.24 | 83.75 | 89.38 | 86.56 | 89.55 | **90.91** | 81.33 |

Table 7: Experimental results on the Potsdam dataset; the values in **bold** are the best.

| Method | Imp. Surf. | Build. | Low veg. | Tree | Car | mean F1(%) | OA(%) | mIoU(%) |
|---|---|---|---|---|---|---|---|---|
| UZ 1 [59] | 89.30 | 95.40 | 81.80 | 80.50 | 86.50 | 86.70 | 85.80 | - |
| FCNs [8] | 89.05 | 93.34 | 83.54 | 83.67 | 89.48 | 87.82 | 86.40 | 78.48 |
| DeepLabv3 [55] | 89.90 | 94.58 | 83.58 | 85.48 | 73.24 | 85.36 | 87.73 | 75.12 |
| S-RA-FCN [13] | 91.33 | 94.70 | 86.81 | 83.47 | 94.52 | 90.17 | 88.59 | 82.38 |

| Method | | | | | | | | |
|---|---|---|---|---|---|---|---|---|
| DeepLabv3+ [46] | 92.27 | 95.52 | 85.71 | 86.04 | 89.42 | 89.79 | 89.60 | 81.69 |
| DST 6 [64] | 92.40 | 96.40 | 86.80 | 87.70 | 93.40 | 91.34 | 90.20 | - |
| DANet [21] | 91.50 | 95.83 | 87.21 | **88.79** | 95.16 | 91.70 | 90.56 | 83.77 |
| AZ3 | 93.10 | 96.30 | 87.20 | 88.60 | 96.00 | 92.24 | 90.70 | - |
| CASIA3 [4] | 93.40 | 96.80 | 87.60 | 88.30 | 96.10 | **92.44** | 91.00 | - |
| PSPNet [62] | 93.36 | **96.97** | **87.75** | 88.50 | 95.42 | 92.40 | 91.08 | 84.88 |
| Heavy-weight Spatial FFPENet | **93.61** | 96.70 | 87.31 | 88.11 | **96.46** | **92.44** | **91.10** | **86.20** |

Figure 8: Qualitative comparisons between our method and the baseline (Deeplabv3+) on the Vaihingen dataset with 512 × 512 patches.

*Comparison with existing methods.*

To evaluate the effectiveness of the segmentation model, we compare our model with other leading benchmark models and the results are shown in Table 6. Specifically, FCNs [8] connect multiscale features by the skip architecture. DeepLabv3 [55] adopts the ASPP module with global pooling operation to capture contextual features. UZ 1 [59] is a CNN model based on encoder–decoder. DeepLabv3+ [46] fuses multiscale features by introducing low-level features to refine high-level features based on DeepLabv3 [55]. S-RA-FCN [13] produces relation-augmented feature representations by the spatial and channel relation modules. ONE 7 [60] fuses the output of two multiscale SegNets [63]. DANet [21] adaptively integrates local features with their global dependencies by two types of attention modules. GSN5 [9] utilizes entropy as a gate function to select features. DLR 10 [61] combines boundary detection with SegNet and FCN. PSPNet [62] exploits the capability of global context information by the pyramid pooling module. Importantly, most of the models adopt the ResNet-101 as the backbone.

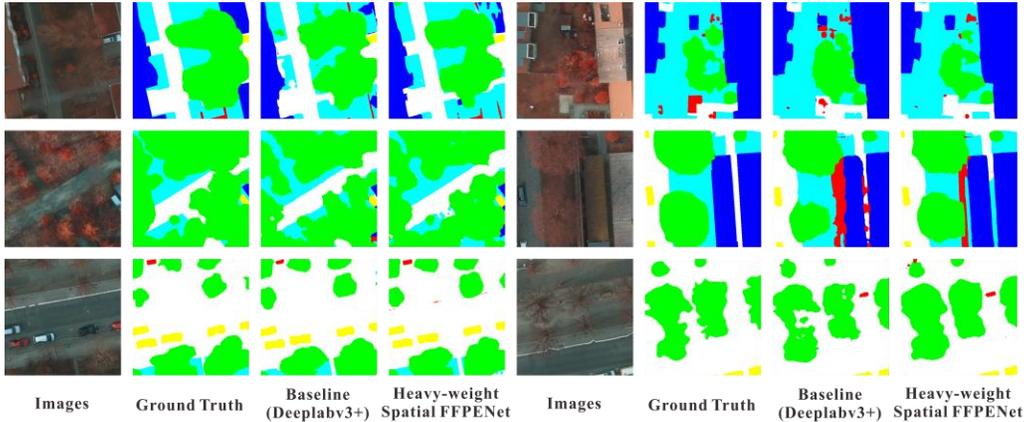

Figure 9: Qualitative comparisons between our method and the baseline (Deeplabv3+) on the Potsdam dataset with 512 × 512 patches.

Table 6 indicates that the heavy-weight spatial FFPNet outperforms other context aggregation or attention-based models in terms of OA. Specifically, we can see that the heavy-weight spatial FFPNet outperforms the baseline model (DeepLabv3+ [46]), with 2.82% and 8.5% increase in OA and mIoU, respectively. Importantly, the qualitative comparisons between our proposed model and the baseline model are provided in Fig. 8. The quantitative and qualitative analyses indicate that our method outperforms the DeepLabv3+ [46] method by a large margin. Furthermore, compared with PSPNet [62], the heavy-weight spatial FFPNet achieves 0.06% improvement in OA but slightly inferior results in some categories such as low vegetation, trees, and cars. However, compared with other high-performance models (such as HMANet [27]) on the Vaihingen dataset, the performance of the heavy-weight spatial FFPNet can be further improved by adopting some strategies such as data augmentation and left-right flipping counterparts during inference.

*3.3.3. Experiments on the Potsdam dataset*

In order to further validate the effectiveness of the segmentation model, we conducted experiments on the Potsdam dataset. Table 7 shows the result of comparison of the proposed model with other excellent models, including DAT 6 [64], an end-to-end self-cascaded network CASIA3 [4], and the other methods used for the comparison on the Vaihingen dataset. Remarkably, the heavy-weight spatial FFPNet achieves an OA of 92.44% and mIoU of 86.20%, which are 0.02% and 1.32% improvement, respectively, compared to the values achieved by PSPNet. In addition, our F1 score of car is much higher than that achieved by PSPNet and exceeds the second best value achieved by CCNet by 1.03%. Thus, the effectiveness of our proposed model in handling multiscale feature fusion for the segmentation of remote sensing images is demonstrated.

In addition, the quantitative comparison results are shown in Fig. 9. The third and fourth columns represent the results of the baseline and the proposed models, respectively. Obviously, the visualization results in the fourth column are better than those in the third column. Moreover, as Table 7 indicates, the proposed model shows an improvement of 1.5% in OA and 4.51% in mIoU compared with the values achieved by DeepLabv3+ [46]. Therefore, it has been further demonstrated that the heavy-weight spatial FFPNet can effectively extract and fuse the spatial features of remote sensing images, thereby improving the segmentation performance of high-resolution images.

### 3.4. Spatial–spectral FFPNet Evaluation on Hyperspectral Datasets

#### 3.4.1. Implementation details

**Data preprocessing.**

When the hyperspectral images are divided into training and testing sets, the imbalance between categories brings difficulties for model training. For example, in the IP dataset, the "Oats" class has 20 labeled pixels, while the "Soybean-mintill" class has 2455 labeled pixels. To ensure comparability of data as much as possible, the same data processing strategy as the baseline [38] is used to deal with the imbalance of categories, that is, optimally selecting the number of samples of each category. Importantly, some lower data setups were added to highlight the superiority of the spatial– spectral FFPNet. Specifically, a maximum number of samples per category was set as a threshold to select the number of samples, that is, 50, 100, 150, and 200, per category. For the richer class samples, we simplified split the samples on the basis of the threshold. On contrary, when the number of class samples was less than twice the threshold, 50% samples of the corresponding class were selected. Detailed training sample schemes for the IP dataset are listed in Table 8, and the same schemes are adopted for the UP dataset as shown in Table 9. It is worth noting that the number of samples in both datasets is less than or equal to that in the baseline [38]; we conducted more sampling schemes for subsequent experiments.

Furthermore, for better numerical optimization, the normalization (obtaining a zero mean and unit variance) can be performed globally on the entire hyperspectral image; this strategy works especially well with the fully connected CNN classifier [5]. Then, a band-mean normalization process was conducted in which every spectral band is normalized by subtracting the mean.

*Model setup.* the proposed spatial–spectral FFPNet was initialized with two strategies: the backbone of the light-weight spatial FFP module was initialized with VGGNet16 pretrained on ImageNet [47]. Importantly, the parameters (weights and bias) of the first convolutional layer in the pretrained network only include three channels, while the hyperspectral classification task requires $p$-channel inputs (for example, 200 channels for the IP dataset and 103 channels for the UP dataset). Therefore, we copied the initialization parameters of the first convolutional layer in the pretrained network until the $p$-channel inputs were reached, similarly to CoinNet [65]. By contrast, the spectral FFP module was initialized with Kaiming uniform distribution. In addition, different from high-resolution experiments, the cross-entropy loss function was used to minimize the spatial–spectral FFPNet parameters because of the quantity limitation of labeled hyperspectral images. Batch normalization and ReLU were used in all layers, except for the classifier layer. Adam [66] was employed as the optimizer with a learning rate of 0.001. The mini-batch size was set to 24 for both datasets, and the maximum epoch was 200. The experiments were conducted on a high-performance computing cluster with one NVIDIA Titan RTX 24GB GPU.

*3.4.2. Experiments on hyperspectral datasets*
***Ablation study.***

In order to analyze the effectiveness of the proposed spatial– spectral FFPNet in hyperspectral image classification, two aspects are mainly considered. First, the size of the training set (number of samples and sample pixel size) is closely related to the deep learning algorithms used in hyperspectral image classification tasks. In addition, data augmentation is widely used in the hyperspectral training set to better solve the optimization problem of many parameters in CNNs. For instance, Chen et al. [39] proposed a virtual sample enhanced method to further improve the performance of CNNs. Thus, to analyze how the training set, including the number of sam- ples and sample patch size, and data augmentation affect the performance of hyperspectral image classification, we conducted different ablation studies. Second, in order to analyze the impact of spatial and spectral models on the performance of hyperspectral image classification, the ablation experiments of the spatial FFPNet, spectral FFPNet, and spatial–spectral FFPNet were conducted for the IP and UP hyperspectral image datasets.

Table 8: Number of training samples used by the spatial–spectral FFPNet for the IP dataset.

| Class | Pixels | 200 samples per category | 150 samples per category | 100 samples per category | 50 samples per category | 200 samples per category in [38] |
|---|---|---|---|---|---|---|
| Alfalfa | 46 | 23 | 23 | 23 | 23 | 33 |
| Corn-notill | 1428 | 200 | 150 | 100 | 50 | 200 |
| Corn-mintill | 830 | 200 | 150 | 100 | 50 | 200 |
| Corn | 237 | 118 | 118 | 100 | 50 | 181 |
| Grass-pasture | 483 | 200 | 150 | 100 | 50 | 200 |
| Grass-trees | 730 | 200 | 150 | 100 | 50 | 200 |
| Grass-pasture-mowed | 28 | 14 | 14 | 14 | 14 | 20 |
| Hay-windrowed | 478 | 200 | 150 | 100 | 50 | 200 |
| Oats | 20 | 10 | 10 | 10 | 10 | 14 |
| Soybeans-notill | 972 | 200 | 150 | 100 | 50 | 200 |
| Soybeans-mintill | 2455 | 200 | 150 | 100 | 50 | 200 |
| Soybeans-clean | 593 | 200 | 150 | 100 | 50 | 200 |
| Wheat | 205 | 102 | 102 | 100 | 50 | 143 |
| Woods | 1265 | 200 | 150 | 100 | 50 | 200 |
| Bldg-grass-tree-drives | 386 | 193 | 150 | 100 | 50 | 200 |
| Stone-steel-towers | 93 | 46 | 46 | 46 | 46 | 75 |
| Total | 10249 | 2306 | 1813 | 1293 | 693 | 2466 |

Table 9: Number of training samples used by the spatial–spectral FFPNet for the UP dataset.

| Class | Pixels | 200 samples per category | 150 samples per category | 100 samples per category | 50 samples per category | 200 samples per category in [38] |
|---|---|---|---|---|---|---|
| Asphalt | 6631 | 200 | 150 | 100 | 50 | 200 |
| Meadows | 18649 | 200 | 150 | 100 | 50 | 200 |
| Gravel | 2099 | 200 | 150 | 100 | 50 | 200 |
| Trees | 3064 | 200 | 150 | 100 | 50 | 200 |
| Painted metal sheets | 1345 | 200 | 150 | 100 | 50 | 200 |
| Bare soil | 5029 | 200 | 150 | 100 | 50 | 200 |
| Bitumen | 1330 | 200 | 150 | 100 | 50 | 200 |
| Self-blocking bricks | 3682 | 200 | 150 | 100 | 50 | 200 |
| Shadows | 947 | 200 | 150 | 100 | 50 | 200 |
| Total | 42776 | 1800 | 1350 | 900 | 450 | 1800 |

**(1) Sample patch size:** we conducted an ablation study for different sample patch sizes. For the IP dataset, patch sizes d = 9, 15, 19 and 29 were considered, and for the UP dataset, d = 9, 15, 21, and 27 were tested. The different utilized for hyperspectral image classification. Table 10 shows the total training times and accuracy results with different patch sizes for a fixed number of samples (100) per category.

Table 10: Total training time (in minutes) and accuracy evaluation with different patch sizes d = 9, 15, 19, and 29 for the IP dataset and d = 9, 15, 21, and 27 for the UP dataset; the values in **bold** are the best. All results are the average of three runs with 100 samples per category; maximum epoch = 200 and mini-batch size = 24.

| Dataset | Patch size | | OA | AA | Kappa |
|---|---|---|---|---|---|
| IP | d = 9 | 15.13 | 96.30 | 98.31 | 95.73 |
|    | d = 15 | 17.86 | 96.78 | 98.68 | 96.29 |
|    | d = 19 | **14.66** | 98.50 | 99.16 | 98.27 |
|    | d = 29 | 21.57 | **98.74** | **99.43** | **98.57** |
| UP | d = 9 | 9.12 | 91.37 | 91.28 | 88.60 |
|    | d = 15 | **7.06** | 96.41 | 96.14 | 95.25 |
|    | d = 21 | 8.16 | **98.82** | **98.37** | **98.44** |
|    | d = 27 | 9.60 | 97.29 | 97.26 | 96.42 |

On both datasets, as more pixels are added, more useful contextual spatial information could be utilized by the spatial–spectral FFPNet model. Thus, the model achieves better performance in the case of more spatial information while also spending more training time. However, as the patch size is further increased, such as d = 27 for the UP dataset, the model performance slightly decreases; this is because a patch containing too many other classes can detract from the target pixel. Specifically, for the IP data, d = 29 obtains the best performance, with an OA of 98.76%. However, the average training time for d = 29 is considerably longer than that for the other groups (almost twice that for d = 19). In terms of the accuracy to time ratio, d = 19 yields the best performance for the IP dataset, with an OA of 98.50% for an average training time of 14.66 minutes. For the UP dataset, d = 21 achieves the best result in terms of the accuracy–time ratio, resulting in an OA of 98.82% for an average training time of 8.16 minutes. In addition, d = 15 requires the minimum training time (7.06 minutes) to achieve an acceptable accuracy (96.41%).

Table 11: Classification accuracies obtained by the proposed spatial–spectral FFPNet (with sample patch sizes d = 9, 15, 19, and 29) for the IP dataset. All results are the average of three runs with maximum epoch = 200 and mini-batch size = 24.

| Sample patch size | d = 9 | | | | d = 15 | | | | d = 19 | | | | d = 29 | | | |
|---|---|---|---|---|---|---|---|---|---|---|---|---|---|---|---|---|
| Samples per category | 50 | 100 | 150 | 200 | 50 | 100 | 150 | 200 | 50 | 100 | 150 | 200 | 50 | 100 | 150 | 200 |
| Alfalfa | **100.00** | **100.00** | 95.65 | 95.65 | **100.00** | **100.00** | 95.65 | **100.00** | **100.00** | **100.00** | **100.00** | **100.00** | **100.00** | **100.00** | **100.00** | **100.00** |
| Corn-notill | 73.66 | 95.63 | **98.83** | 98.78 | 74.38 | 94.65 | 98.51 | **99.35** | 83.16 | 99.32 | **99.53** | 99.59 | 87.11 | 99.72 | 100.00 | 100.00 |
| Corn-mintill | 72.69 | 98.63 | **98.97** | 99.84 | 86.03 | 97.40 | **99.71** | 98.25 | 97.44 | 96.30 | **99.56** | 99.05 | 95.65 | 100.00 | 100.00 | 100.00 |
| Corn | 93.58 | 100.00 | 100.00 | 100.00 | 100.00 | 100.00 | 100.00 | 100.00 | 98.93 | 100.00 | 100.00 | 100.00 | 100.00 | 100.00 | 100.00 | 100.00 |
| Grass-pasture | 90.76 | 98.96 | 98.80 | 100.00 | 97.46 | 99.48 | 100.00 | 100.00 | 93.07 | 97.65 | 98.80 | **98.94** | 95.83 | 98.33 | 100.00 | 100.00 |
| Grass-trees | 97.06 | 98.10 | 99.31 | **99.62** | 97.50 | 99.84 | 100.00 | 100.00 | 97.35 | 99.05 | 100.00 | 98.87 | 95.05 | 98.35 | 99.45 | 100.00 |
| Grass-pasture-mowed | 100.00 | 100.00 | 100.00 | 100.00 | 100.00 | 100.00 | 100.00 | 100.00 | 100.00 | 100.00 | 100.00 | 100.00 | 100.00 | 100.00 | 100.00 | 100.00 |
| Hay-windrowed | 97.20 | **100.00** | **100.00** | **100.00** | 98.36 | 100.00 | 100.00 | 100.00 | 100.00 | 100.00 | 100.00 | 100.00 | 100.00 | 100.00 | 100.00 | 100.00 |
| Oats | **100.00** | **100.00** | **100.00** | **100.00** | **100.00** | **100.00** | **100.00** | **100.00** | **100.00** | **100.00** | **100.00** | **100.00** | **100.00** | **100.00** | **100.00** | **100.00** |
| Soybeans-notill | 78.98 | 94.01 | **99.75** | 99.33 | 87.04 | 99.19 | 97.78 | **99.73** | 87.15 | 97.81 | 99.51 | 100.00 | 98.26 | 98.70 | 99.57 | 100.00 |
| Soybeans-mintill | 70.73 | 93.25 | 94.84 | **98.58** | 74.80 | 93.25 | 97.53 | **99.38** | 80.79 | 97.54 | 95.75 | **99.78** | 91.19 | 96.41 | **99.35** | **99.35** |
| Soybean-clean | 86.37 | 99.19 | **99.55** | 99.24 | 94.66 | 98.58 | 98.87 | **100.00** | 96.87 | 98.99 | **100.00** | **100.00** | 98.65 | 99.32 | 99.32 | 100.00 |
| Wheat | 98.71 | 100.00 | 100.00 | 100.00 | 100.00 | 100.00 | 100.00 | 100.00 | 100.00 | 100.00 | 100.00 | 100.00 | 100.00 | 100.00 | 100.00 | 100.00 |
| Woods | 94.07 | 97.94 | 98.57 | **98.78** | 96.79 | 98.63 | **99.64** | 99.34 | 93.83 | 99.91 | 100.00 | 100.00 | 98.73 | 100.00 | 99.68 | 100.00 |
| Bldg-grass-tree-drives | 99.11 | 97.20 | 100.00 | 100.00 | 98.51 | 97.90 | 100.00 | 100.00 | 100.00 | 100.00 | 100.00 | 100.00 | 96.88 | 100.00 | 100.00 | 100.00 |
| Stone-steel-towers | **100.00** | **100.00** | **100.00** | 95.74 | **100.00** | **100.00** | **100.00** | **100.00** | 97.87 | 100.00 | 100.00 | 100.00 | 100.00 | 100.00 | 100.00 | 100.00 |
| OA | 82.12 | 96.30 | 97.98 | **99.07** | 86.44 | 96.78 | 98.74 | **99.47** | 89.79 | 98.50 | 98.63 | **99.68** | 94.65 | 98.74 | 99.69 | **99.84** |
| AA | 90.81 | 98.31 | 99.02 | **99.10** | 94.10 | 98.68 | 99.23 | **99.75** | 95.40 | 99.16 | 99.57 | **99.76** | 97.34 | 99.43 | 99.84 | **99.96** |
| Kappa | 79.65 | 95.73 | 97.65 | **98.90** | 84.56 | 96.29 | 98.53 | **99.38** | 88.35 | 98.27 | 98.41 | **99.63** | 93.93 | 98.57 | 99.64 | **99.82** |
| Run time | **9.24** | 15.13 | 20.39 | 22.55 | **8.21** | 17.86 | 28.47 | 31.44 | **10.37** | 14.66 | 23.12 | 34.05 | **15.43** | 21.57 | 27.41 | 37.38 |

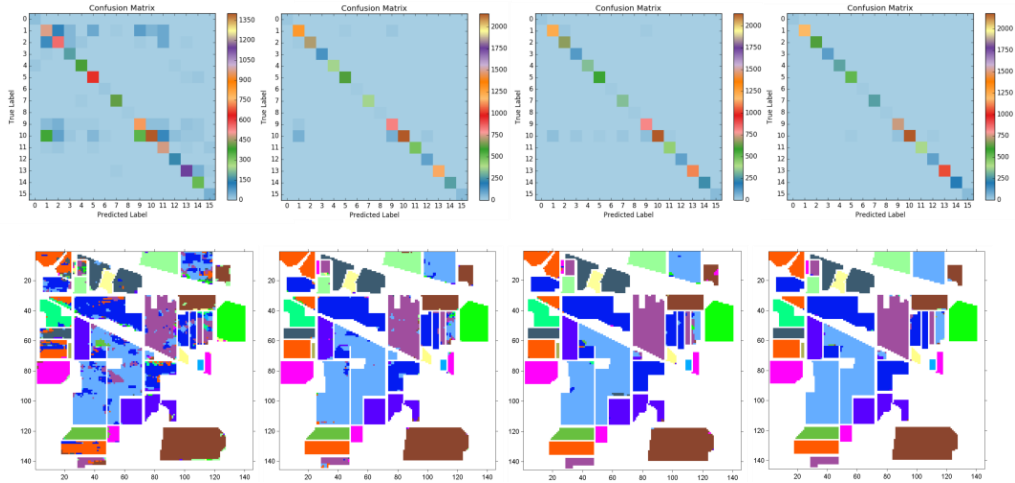

Figure 10: Classification results for the IP image with d = 9 and number of samples per category = 50 (first column), 100 (second column), 150 (third column), and 200 (fourth column). The upper row represents the visualization results of the confusion matrix for each category on the testing set (the more prominent the color of the diagonal area, the better the result) and the lower row represents the qualitative results of classification

**(2) Sample per category:** in order to evaluate the impact of the number of samples per category on the model performance, many experiments with different patch sizes and different number of training samples, that is, 50, 100, 150, and 200, per category were conducted.

The classification accuracy results in terms of OA, AA, and kappa coefficients obtained for the IP dataset are presented in Table 11.

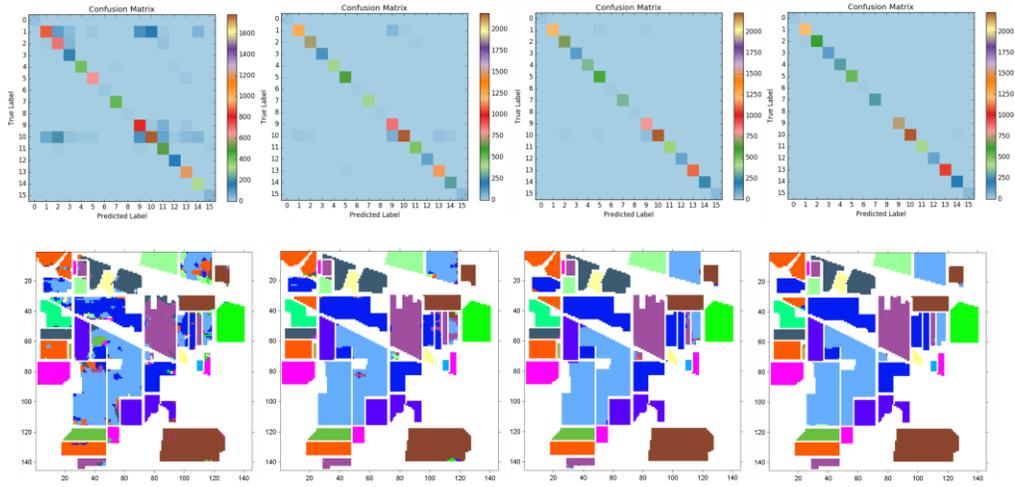

Figure 11: Classification results for the IP image with d = 15 and number of samples per category = 50 (first column), 100 (second column), 150 (third column), and 200 (fourth column). The upper row represents the visualization results of the confusion matrix for each category on the testing set (the more prominent the color of the diagonal area, the better the result) and the lower row represents the qualitative results of classification

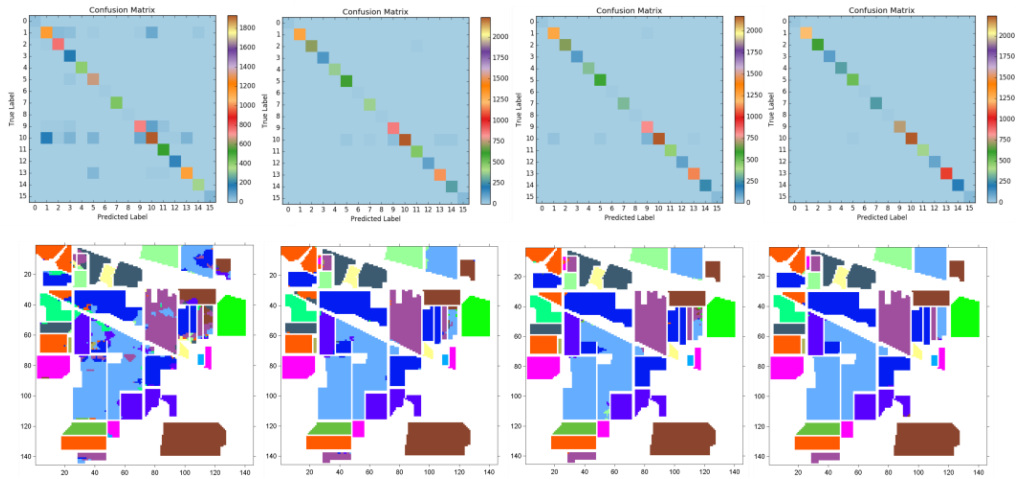

Figure 12: Classification results for the IP image with d = 19 and number of samples per category = 50 (first column), 100 (second column), 150 (third column), and 200 (fourth column). The upper row represents the visualization results of the confusion matrix for each category on the testing set (the more prominent the color of the diagonal area, the better the result) and the lower row represents the qualitative results of classification

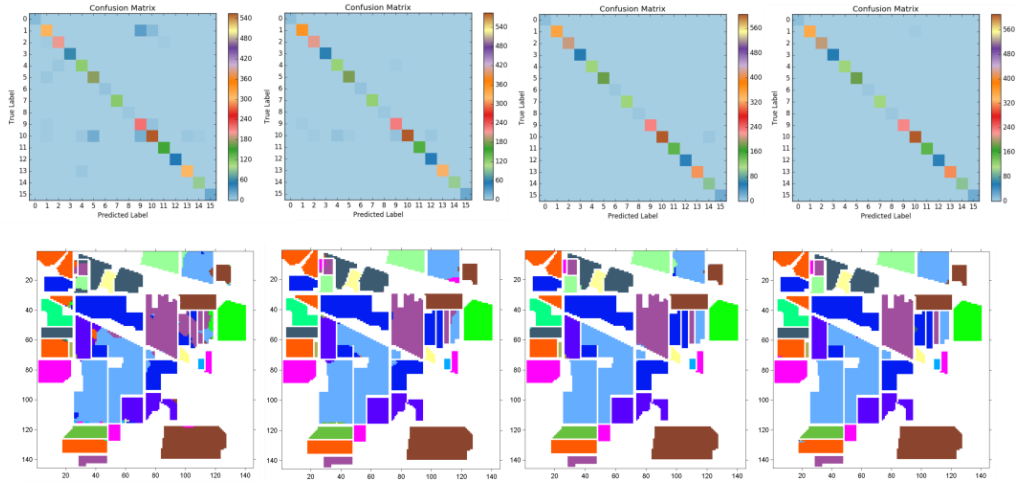

Figure 13: Classification results for the IP image with d = 29 and number of samples per category = 50 (first column), 100 (second column), 150 (third column), and 200 (fourth column). The upper row represents the visualization results of the confusion matrix for each category on the testing set (the more prominent the color of the diagonal area, the better the result) and the lower row represents the qualitative results of classification

Obviously, according to the results of each patch size (d = 9, 15, 19, and 29), as more training samples per category are added, the accuracy of the proposed model classification increases, and the training time also increases. Concretely, when the number of samples is small, the model can achieve a superior classification result; that is, with d = 9, 15, 19, and 29 and 50 samples per category, OA values of 82.12%, 86.44%, 89.79%, and 94.64%, respectively, are achieved. Therefore, it is confirmed that the spatial–spectral FFPNet model can fully use multiscale spatial and spectral information to achieve more robust and accurate end-to-end hyperspectral image classification with a small number of training samples. Furthermore, with 200 samples per category, the best OA value of 99.84% of all groups is achieved for d = 29, and the OA values for d = 9, 15, and 19 vary by not more than 0.8%. All of the groups with 200 samples per category attain values above 99%; this further shows the spatial–spectral FFPNets robustness and the ability to express and extract multiscale features.

The qualitative results obtained for the IP dataset for patch sizes of d = 9, 15, 19, and 29, respectively, with 50, 100, 150, and 200 samples per category, are provided in Fig. 10–13. First, the visualization results of the confusion matrix for each category indicate that as more training samples are added, the color of the diagonal area gets brighter, while the other areas become more unified to blue.

This indicates that the classification results of each class are improving. In addition, as the patch size increases, the accuracy of each class increases. However, when relatively adequate training samples are used in the network, the accuracy of each class is relatively similar (e.g., d = 9, 15, 19, and 29 with 200 samples per category, and d = 29 with 150 samples per category). Second, according to the classification maps acquired from each experiment, shown in Fig. 10–13, the best results are achieved with 200 samples per category, especially for d = 19 and 29 with 200 samples per category; these are the most similar to the ground truth map of the IP image. Specifically, when the number of spatial pixels is small (d = 9, 15, and 19 with 50 samples per category), a small part of the middle pixels of the areas in some categories could be misclassified, especially for d = 9 with 50 samples per category. However, when the number of training samples per category increases to 100, these middle pixels are accurately classified. Furthermore, when the number of training samples is less, a small number of pixels near the edges are easily misclassified; we call this the "*boundary error effect*". However, with increasing training samples, the boundary error effect gradually weakens or even disappears for 150 training samples per category. More importantly, the overall classification result is generally excellent when the sample size is extremely small (i.e., for d = 9, 15, 19, and 29 with 50 samples per category). This demonstrates that the proposed model can better address the problem of overfitting when less hyperspectral samples are available.

Table 12: Classification accuracies obtained by the proposed spatial–spectral FFPNet (with sample patch sizes of d = 9, 15, 21, and 27) for the UP dataset. All results are the average of three runs with maximum epoch = 200 and mini-batch size = 24.

| Sample patch size | d = 9 | | | | d = 15 | | | | d = 21 | | | | d = 27 | | | |
|---|---|---|---|---|---|---|---|---|---|---|---|---|---|---|---|---|
| Samples per category | 50 | 100 | 150 | 200 | 50 | 100 | 150 | 200 | 50 | 100 | 150 | 200 | 50 | 100 | 150 | 200 |
| Asphalt | 66.14 | 88.65 | 95.72 | **96.32** | 78.03 | 94.57 | **98.19** | **98.19** | 93.54 | 97.34 | **98.97** | 97.59 | 92.28 | 91.61 | 99.70 | **99.82** |
| Meadows | 87.47 | 94.64 | **98.09** | 97.96 | 92.86 | 97.98 | 99.44 | **99.98** | 97.68 | 99.81 | 99.89 | **99.91** | 97.45 | 99.31 | 99.68 | **99.85** |
| Gravel | 49.62 | 90.27 | **97.14** | 95.99 | 91.03 | 96.37 | **99.43** | 98.66 | 89.12 | 98.09 | **99.81** | **99.81** | 93.32 | 98.66 | **99.05** | **99.05** |
| Trees | 57.44 | 86.03 | 87.86 | **90.60** | 59.79 | 92.95 | 96.08 | **96.61** | 72.19 | 94.39 | 94.39 | **96.34** | 83.16 | 93.34 | 94.26 | **96.61** |
| Painted metal sheets | 95.24 | 97.62 | 99.40 | **100.00** | 97.92 | 98.81 | **99.70** | 99.11 | 98.51 | 98.51 | **100.00** | **100.00** | 96.13 | 99.40 | 99.70 | **100.00** |
| Bare Soil | 18.46 | 87.59 | **96.26** | 93.95 | 60.14 | 98.41 | 99.68 | **100.00** | 94.99 | 99.68 | **100.00** | **100.00** | 95.78 | **100.00** | **100.00** | **100.00** |
| Bitumen | 94.88 | 98.49 | **99.40** | 97.59 | 98.49 | 98.80 | 99.40 | **100.00** | 97.59 | **100.00** | **100.00** | **100.00** | 97.59 | **100.00** | **100.00** | **100.00** |
| Self-Blocking Bricks | 36.96 | 84.46 | 95.98 | **96.30** | 36.74 | 90.00 | 98.70 | **99.02** | 77.39 | **99.24** | 97.28 | 98.04 | 71.09 | 93.91 | 97.72 | **99.13** |
| Shadows | 79.65 | 93.81 | 96.02 | **97.35** | 79.65 | 97.35 | **100.00** | 98.23 | 93.36 | 98.23 | 98.67 | **100.00** | 95.13 | 99.12 | 95.58 | **99.56** |
| OA | 67.99 | 91.37 | **96.58** | 96.51 | 79.47 | 96.41 | 98.99 | **99.25** | 92.66 | 98.82 | 99.12 | **99.15** | 92.87 | 97.29 | 99.05 | **99.53** |
| AA | 65.09 | 91.28 | 96.21 | **96.23** | 77.18 | 96.14 | **98.96** | 98.87 | 90.49 | 98.37 | 98.78 | **99.08** | 91.33 | 97.26 | 98.41 | **99.33** |
| Kappa | 56.52 | 88.60 | **95.47** | 95.37 | 72.73 | 95.25 | 98.66 | **99.01** | 90.24 | 98.44 | 98.83 | **98.87** | 90.55 | 96.42 | 98.75 | **99.38** |
| Run time | 4.91 | 9.12 | 13.23 | 15.48 | **4.79** | 7.06 | 12.40 | 14.05 | **4.90** | 8.16 | 11.50 | 14.55 | **5.04** | 9.60 | 13.66 | 18.45 |

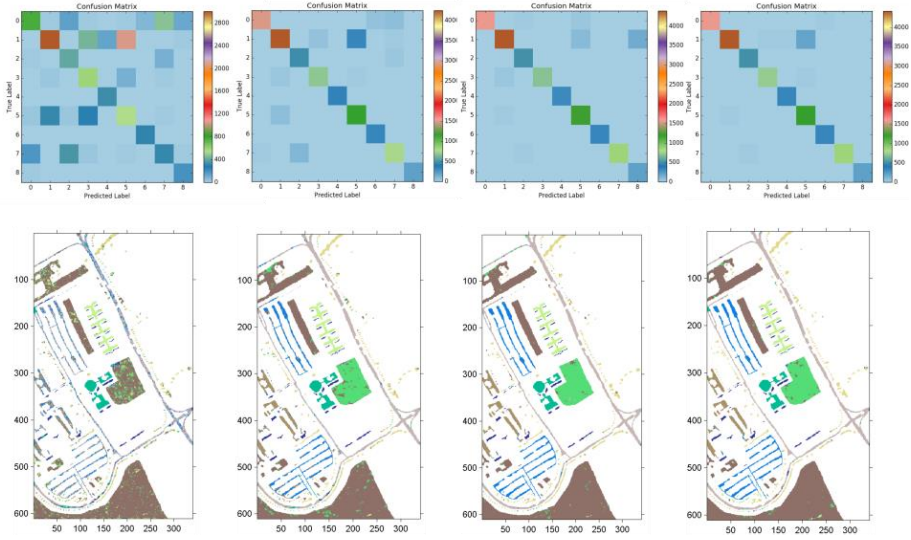

Figure 14: Classification results for the UP image with d = 9 and number of samples per category = 50 (first column), 100 (second column), 150 (third column), and 200 (fourth column). The upper row represents the visualization results of the confusion matrix for each category on the testing set (the more prominent the color of the diagonal area, the better the result) and the lower row represents the qualitative results of classification

Table 12 lists the results for the UP dataset. For every patch size, as the training samples increase, the model performance gradually improves. Notably, the model performance is more sensitive when the sample size is exceedingly small. For example, for d = 9 and 15 with 50 samples per category, the model performance in terms of OA for the UP dataset is less than 80%, while it improves to more than 90% when the samples per category are increased to 100. Furthermore, the sensitivity of the model performance to the small amount of training data and large number of model parameters can be better addressed by *data enhancement methods* such as random rotation and addition of random noise. The effectiveness of data augmentation will be discussed in the third ablation analysis. Furthermore, d = 27 with 200 samples per category can be regarded as the best setting for the UP dataset as an OA of 99.53% is achieved. However, the model performance for d = 15 and 21 with 200 samples per category differs from the performance for d = 27 with 200 samples per category by less than 0.4%. In terms of the training time, d = 15 or d = 21 is a better choice.

The qualitative classification results obtained for the UP dataset for patch

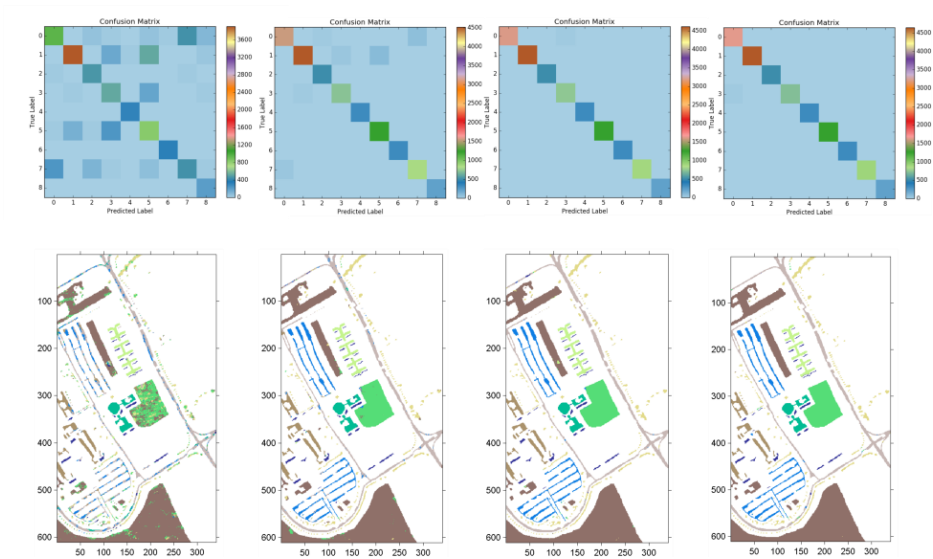

Figure 15: Classification results for the UP image with d = 15 and number of samples per category = 50 (first column), 100 (second column), 150 (third column), and 200 (fourth column). The upper row represents the visualization results of the confusion matrix for each category on the testing set (the more prominent the color of the diagonal area, the better the result) and the lower row represents the qualitative results of classification

sizes d = 9, 15, 21, and 27, respectively, with 50, 100, 150 and 200 samples per category are provided in Fig. 14–17. First, according to the visualization results of the confusion matrix, as the number of samples increases, the accuracy of every class gradually improves. Specifically, when the samples are limited (i.e., d = 9 and 15 with 50 samples per category), the classification accuracies for some classes are higher, such as Meadows, Painted metal sheets Bitumen, and Shadows. When the training samples are adequate (with d = 21 and 100 samples per category for the UP dataset), the classification accuracy of the model in each class is superior (at least 90%). Second, according to the classification maps obtained from each experiment, as more samples are added, the classification results are better, and the configuration of 200 samples per category acquires the best results in each group. Compared with other groups, the middle pixels of the areas in some class are easily misidentified and a small number of pixels near the border are affected by the boundary error effect, as shown in Fig. 14. These misidentified areas improve as the number of samples and patch size increase. Importantly, regardless of the number of samples, the visual classification results for the

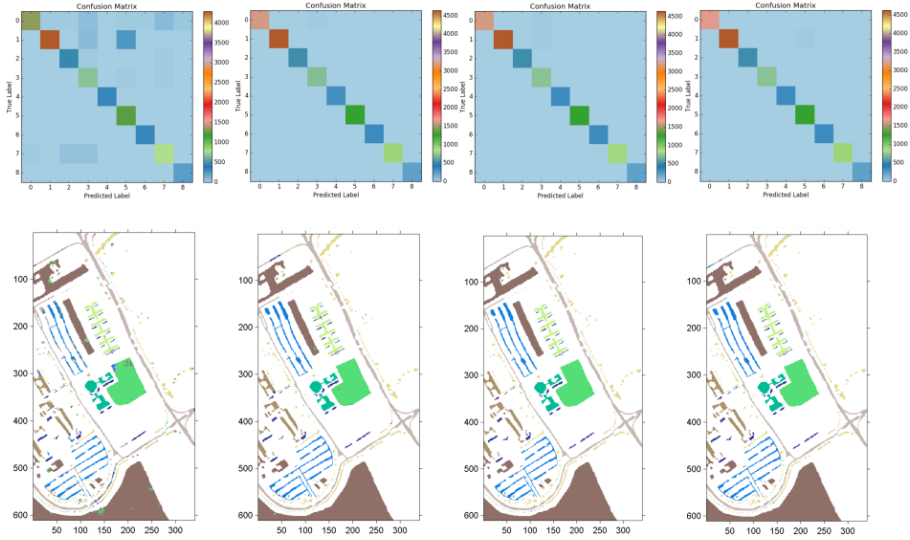

Figure 16: Classification results for the UP image with d = 21 and number of samples per category = 50 (first column), 100 (second column), 150 (third column), and 200 (fourth column). The upper row represents the visualization results of the confusion matrix for each category on the testing set (the more prominent the color of the diagonal area, the better the result) and the lower row represents the qualitative results of classification

UP dataset are favorable. It has been further demonstrated that the spatial– spectral FFPNet has greater robustness to deal with the challenge of a small number of hyperspectral labeled samples and the ability to express spatial and spectral features of hyperspectral image data.

**(3) Data enhancement:** we used random horizontal and vertical flips, random rotation (with angles 90°, 180°, and 270°) to enhance small-scale hyperspectral datasets. To further test the effectiveness of the spatial–spectral FFPNet subjected to data enhancement when the training samples are intensely limited, 50 samples per category for the IP dataset and the UP dataset were utilized. To highlight the superiority of the proposed model purely, it is worth noting that data enhancement techniques were not used in other experiments in this paper because the baseline model [38] does not use data enhancement.

The ablation study results of data enhancement, with sample patch sizes of d = 9, 15, 19, and 29 and 50 samples per category on the IP and UP datasets are shown in Table 13 and Table 14, respectively. Clearly, the classification accuracy of the spatial–spectral FFPNet with data enhancement

Table 13: Ablation study of data enhancement to spatial–spectral FFPNet performance (with sample patch sizes of d = 9, 15, 19, and 29 and 50 samples per category) on the IP dataset; the values in **bold** are the best. All results are the average of three runs with maximum epoch = 200 and mini-batch size = 24.

| Methods | d = 9 | | d = 15 | | d = 19 | | d = 29 | |
|---|---|---|---|---|---|---|---|---|
| Class | Spatial–spectral FFPNet | Spatial–spectral FFPNet + Data enhancement | Spatial–spectral FFPNet | Spatial–spectral FFPNet + Data enhancement | Spatial–spectral FFPNet | Spatial–spectral FFPNet + Data enhancement | Spatial–spectral FFPNet | Spatial–spectral FFPNet + Data enhancement |
| Alfalfa | **100.00** | **100.00** | **100.00** | **100.00** | **100.00** | **100.00** | **100.00** | **100.00** |
| Corn-notill | 73.66 | **82.22** | 74.38 | **89.84** | 83.16 | **96.81** | 87.11 | **98.60** |
| Corn-min | 72.69 | **95.90** | 86.03 | **95.64** | 97.44 | **98.46** | 95.65 | **98.55** |
| Corn | 93.58 | **98.40** | **100.00** | **100.00** | 98.93 | **100.00** | **100.00** | 98.31 |
| Grass/Pasture | 90.76 | **91.69** | **97.46** | 97.00 | 93.07 | **94.46** | **95.83** | **95.83** |
| Grass/Trees | **97.06** | 94.12 | 97.50 | **98.68** | 97.35 | **98.09** | 95.05 | **100.00** |
| Grass/pasture-mowed | **100.00** | **100.00** | **100.00** | **100.00** | **100.00** | **100.00** | **100.00** | **100.00** |
| Hay-windrowed | 97.20 | **99.77** | 98.36 | **99.30** | **100.00** | **100.00** | **100.00** | **100.00** |
| Oats | **100.00** | **100.00** | **100.00** | **100.00** | **100.00** | **100.00** | **100.00** | **100.00** |
| Soybeans-notill | 78.98 | **87.25** | **87.04** | 84.53 | 87.15 | **96.08** | **98.26** | 97.39 |
| Soybeans-min | 70.73 | **84.24** | 74.80 | **93.43** | 80.79 | **94.97** | 91.19 | **94.62** |
| Soybean-clean | 86.37 | **97.97** | 94.66 | **99.08** | 96.87 | **97.05** | 98.65 | **99.32** |
| Wheat | 98.71 | **100.00** | **100.00** | **100.00** | **100.00** | **100.00** | **100.00** | **100.00** |
| Woods | 94.07 | **95.06** | 96.79 | **99.92** | 93.83 | **98.60** | 98.73 | **100.00** |
| Bldg-Grass-Tree-Drives | 99.11 | **99.70** | 98.51 | **99.11** | **100.00** | 99.70 | 96.88 | **100.00** |
| Stone-steel | **100.00** | **100.00** | **100.00** | **100.00** | 97.87 | 97.87 | **100.00** | **100.00** |
| OA | 82.12 | **90.32** | 86.44 | **94.68** | 89.79 | **97.02** | 94.65 | **97.88** |
| AA | 90.81 | **95.39** | 94.10 | **97.28** | 95.40 | **98.26** | 97.34 | **98.91** |
| Kappa | 79.65 | **88.95** | 84.56 | **93.89** | 88.35 | **96.58** | 93.93 | **97.58** |

Table 14: Ablation study of data enhancement to spatial–spectral FFPNet performance (with sample patch sizes of d = 9, 15, 21, and 27 and 50 samples per category on the UP dataset; the values in **bold** are the best. All results are the average of three runs with maximum epoch = 200 and mini-batch size = 24.

| Methods | d = 9 | | d = 15 | | d = 21 | | d = 27 | |
|---|---|---|---|---|---|---|---|---|
| Class | Spatial–spectral FFPNet | Spatial–spectral FFPNet + Data enhancement | Spatial–spectral FFPNet | Spatial–spectral FFPNet + Data enhancement | Spatial–spectral FFPNet | Spatial–spectral FFPNet + Data enhancement | Spatial–spectral FFPNet | Spatial–spectral FFPNet + Data enhancement |
| Asphalt | 66.14 | **91.37** | 78.03 | **90.83** | 93.54 | **97.16** | 92.28 | 88.17 |
| Meadows | 87.47 | **89.94** | 92.86 | **98.07** | 97.68 | **98.33** | 97.45 | **97.73** |
| Gravel | 49.62 | **72.90** | **91.03** | 88.17 | 89.12 | **91.22** | 93.32 | **96.95** |
| Trees | 57.44 | **84.73** | 59.79 | **90.08** | 72.19 | **81.07** | 83.16 | **90.73** |
| Painted metal sheets | 95.24 | **96.73** | 97.92 | **98.81** | 98.51 | **99.40** | 96.13 | **98.21** |
| Bare Soil | 18.46 | **85.28** | 60.14 | **99.28** | 94.99 | **97.69** | 95.78 | **98.81** |
| Bitumen | **94.88** | 92.47 | 98.49 | **99.70** | 97.59 | **100.00** | 97.59 | **100.00** |
| Self-Blocking Bricks | 36.96 | **80.00** | 36.74 | **87.72** | 77.39 | **91.74** | 71.09 | **94.35** |
| Shadows | 79.65 | **93.81** | 79.65 | **92.48** | 93.36 | **97.79** | 95.13 | **96.02** |
| OA | 67.99 | **87.92** | 79.47 | **95.09** | 92.66 | **95.99** | 92.87 | **95.59** |
| AA | 65.09 | **87.47** | 77.18 | **93.90** | 90.49 | **94.93** | 91.33 | **95.66** |
| Kappa | 56.52 | **84.19** | 72.73 | **93.52** | 90.24 | **94.68** | 90.55 | **94.19** |

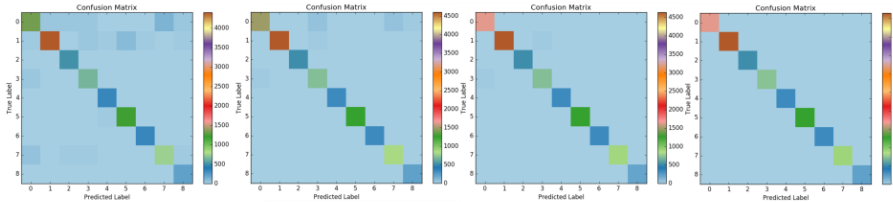

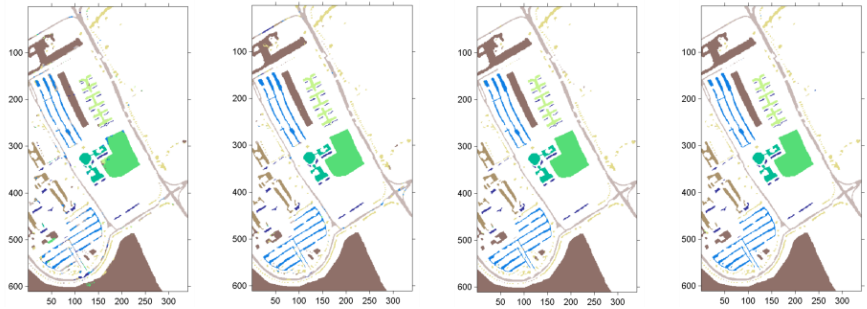

Figure 17: Classification results for the UP image with d = 27 and number of samples per category = 50 (first column), 100 (second column), 150 (third column), and 200 (fourth column). The upper row represents the visualization results of the confusion matrix for each category on the testing set (the more prominent the color of the diagonal area, the better the result) and the lower row represents the qualitative results of classification

is significantly higher than that of the spatial–spectral FFPNet. The accuracy difference between the two datasets reaches 2%–20% in terms of OA. The performance of the model with data enhancement is more dominant for the UP dataset, which is due to the lower ratio of training samples to total samples. Specifically, with d = 9, for the IP dataset, the difference between the spatial–spectral FFPNet with data enhancement and the spatial–spectral FFPNet models is 8.20%, while the corresponding difference for the UP dataset is 19.93%. As the spatial information increases (patch size increases), the performance advantage of the data-enhanced model gradually decreases. For example, in terms of OA, the performance of data-enhanced model is 3.22% higher than that of spatial–spectral FFPNet for the IP dataset with d = 29, and the performance of the spatial–spectral FFPNet improves by 2.71% for the UP dataset with d = 27 by data enhancement. Thus, in case of an extremely small quantity of the labeled hyperspectral dataset, the spatial–spectral FFPNet with data enhancement may be the best choice.

**(4) Spatial FFPNet, spectral FFPNet, and spatial–spectral FFPNet:** As mentioned in the method section, the spectral FFPNet focuses

Table 15: Ablation study of the spatial FFPNet, spectral FFPNet, and spatial–spectral FFPNet with sample patch sizes of d = 9, 15, 19, and 29 and 100 samples per category on the IP dataset; the values in **bold** are the best. All results are the average of three runs with maximum epoch = 200 and mini-batch size = 24.

| Sample patch size | d = 9 | | | d = 15 | | | d = 19 | | | d = 29 | | |
|---|---|---|---|---|---|---|---|---|---|---|---|---|
| Models | Spatial FFPNet | Spectral FFPNet | Spatial–spectral FFPNet | Spatial FFPNet | Spectral FFPNet | Spatial–spectral FFPNet | Spatial FFPNet | Spectral FFPNet | Spatial–spectral FFPNet | Spatial FFPNet | Spectral FFPNet | Spatial–spectral FFPNet |
| Alfalfa | **100.00** | **100.00** | **100.00** | **100.00** | **100.00** | **100.00** | 95.65 | **100.00** | **100.00** | **100.00** | **100.00** | **100.00** |
| Corn-notill | 92.77 | 81.10 | **95.63** | 93.07 | **95.48** | 94.65 | **99.47** | 96.39 | 99.32 | 98.04 | **99.72** | **99.72** |
| Corn-mintill | 96.30 | 95.75 | **98.63** | 92.74 | 96.30 | **97.40** | 96.71 | **97.53** | 96.30 | 99.52 | **100.00** | **100.00** |
| Corn | **100.00** | **100.00** | **100.00** | **100.00** | **100.00** | **100.00** | **100.00** | **100.00** | **100.00** | **100.00** | **100.00** | **100.00** |
| Grass-pasture | 96.61 | 98.43 | **98.96** | 97.13 | 97.13 | **99.48** | 97.91 | 95.82 | 97.65 | 94.17 | 96.67 | **98.33** |
| Grass-trees | 95.87 | 97.94 | **98.10** | 97.78 | 95.56 | **99.84** | 98.10 | 99.52 | **99.05** | **98.90** | 97.80 | 98.35 |
| Grass-pasture-mowed | 57.14 | **100.00** | **100.00** | **100.00** | **100.00** | **100.00** | **100.00** | **100.00** | **100.00** | **100.00** | **100.00** | **100.00** |
| Hay-windrowed | 99.74 | 98.94 | **100.00** | 98.41 | 99.47 | **100.00** | **100.00** | **100.00** | **100.00** | **100.00** | **100.00** | **100.00** |
| Oats | **100.00** | 90.00 | **100.00** | **100.00** | **100.00** | **100.00** | **100.00** | **100.00** | **100.00** | **100.00** | **100.00** | **100.00** |
| Soybeans-notill | 90.09 | 86.87 | **94.01** | 96.43 | 95.28 | **99.19** | 95.70 | 96.31 | **97.81** | 98.26 | 98.26 | **98.70** |
| Soybeans-mintill | 87.90 | 81.44 | **93.25** | **95.16** | 91.00 | 93.25 | 95.54 | 95.12 | **97.54** | **98.04** | 96.90 | 96.41 |
| Soybean-clean | 95.94 | 96.55 | **99.19** | 96.15 | 97.77 | **98.58** | **99.19** | 96.75 | 98.99 | 97.30 | **100.00** | 99.32 |
| Wheat | **100.00** | **100.00** | **100.00** | **100.00** | **100.00** | **100.00** | **100.00** | **100.00** | **100.00** | **100.00** | **100.00** | **100.00** |
| Woods | 95.71 | 95.62 | **97.94** | 98.28 | 98.45 | **98.63** | **99.91** | 99.48 | **99.91** | **100.00** | **100.00** | **100.00** |
| Bldg-grass-tree-drives | **99.30** | 96.50 | 97.20 | 96.50 | 97.20 | **97.90** | **100.00** | **100.00** | **100.00** | **100.00** | 98.96 | **100.00** |
| Stone-steel towers | 97.87 | 95.74 | **100.00** | 97.87 | 95.74 | **100.00** | **100.00** | **100.00** | **100.00** | **100.00** | **100.00** | **100.00** |
| OA | 93.15 | 89.53 | **96.30** | 95.86 | 95.31 | **96.78** | 97.97 | 97.16 | **98.50** | 98.55 | 98.70 | **98.74** |
| AA | 94.08 | 94.68 | **98.31** | 97.47 | 97.46 | **98.68** | 98.76 | 98.56 | **99.16** | 99.01 | 99.27 | **99.43** |
| Kappa | 92.10 | 87.95 | **95.73** | 95.20 | 94.58 | **96.29** | 97.65 | 96.72 | **98.27** | 98.34 | 98.52 | **98.57** |
| Run time | 13.37 | **5.80** | 15.13 | 15.88 | **8.68** | 17.86 | 14.42 | **8.48** | 14.66 | 16.67 | **16.01** | 21.57 |

Table 16: Ablation study of the spatial FFPNet, spectral FFPNet, and spatial–spectral FFPNet with sample patch sizes of d = 9, 15, 21, and 27 and 100 samples per category on the UP dataset; the values in **bold** are the best. All results are the average of three runs with maximum epoch = 200 and mini-batch size = 24.

| Sample patch size | d = 9 | | | d = 15 | | | d = 21 | | | d = 27 | | |
|---|---|---|---|---|---|---|---|---|---|---|---|---|
| Models | Spatial FFPNet | Spectral FFPNet | Spatial–spectral FFPNet | Spatial FFPNet | Spectral FFPNet | Spatial–spectral FFPNet | Spatial FFPNet | Spectral FFPNet | Spatial–spectral FFPNet | Spatial FFPNet | Spectral FFPNet | Spatial–spectral FFPNet |
| Asphalt | 79.48 | 57.82 | **88.65** | 90.04 | 81.41 | **94.57** | 91.43 | 94.39 | **97.34** | 88.47 | 84.67 | **91.61** |
| Meadows | 84.34 | 62.46 | **94.64** | 95.35 | 96.42 | **97.98** | 98.01 | 97.81 | **99.81** | 98.26 | 95.92 | **99.31** |
| Gravel | 76.91 | 68.70 | **90.27** | 90.27 | 91.79 | **96.37** | 98.09 | 94.47 | **98.09** | 97.90 | 97.52 | **98.66** |
| Trees | 76.89 | 54.44 | **86.03** | 80.68 | 78.98 | **92.95** | 87.60 | 69.84 | **94.39** | 93.08 | 83.29 | **93.34** |
| Painted metal sheets | **97.62** | 93.45 | **97.62** | 94.35 | 97.92 | **98.81** | **99.11** | 96.73 | 98.51 | **99.40** | 98.51 | **99.40** |
| Bare SoilC | 60.78 | 29.91 | **87.59** | 90.06 | 89.58 | **98.41** | **99.92** | 99.12 | 99.68 | 99.68 | 99.84 | **100.00** |
| Bitumen | 94.58 | 96.08 | **98.49** | 97.89 | 97.89 | **98.80** | 99.40 | **100.00** | **100.00** | **100.00** | **100.00** | **100.00** |
| Self-Blocking Bricks | 63.15 | 37.28 | **84.46** | 86.30 | 64.13 | **90.00** | 88.70 | 84.67 | **99.24** | 85.65 | 87.61 | **93.91** |
| Shadows | 91.59 | 91.59 | **93.81** | 92.48 | 91.59 | **97.35** | 97.35 | 97.35 | **98.23** | 97.35 | 93.36 | **99.12** |
| OA | 78.98 | 58.11 | **91.37** | 91.81 | 89.02 | **96.41** | 95.73 | 94.16 | **98.82** | 95.51 | 93.25 | **97.29** |
| AA | 80.59 | 65.75 | **91.28** | 90.82 | 87.75 | **96.14** | 95.51 | 92.71 | **98.37** | 95.53 | 93.41 | **97.26** |
| Kappa | 72.42 | 47.10 | **88.60** | 89.23 | 85.50 | **95.25** | 94.37 | 92.22 | **98.44** | 94.07 | 91.09 | **96.42** |
| Run time | 5.83 | **2.24** | 9.12 | 5.04 | **3.12** | 7.06 | 6.57 | **3.39** | 8.16 | 7.63 | **4.97** | 9.60 |

on the extraction and fusion of multi-scale spectral features, while the spatial FFPNet focuses more on effective extraction and integration of context spatial features by the use of attention-based modules. The effects of spectral-only and spatial-only models on the performance of hyperspectral data classification as well as the effectiveness of the spatial–spectral FFPNet model require further analysis. Thus, we conducted ablation studies on the spatial-only, spectral-only, and spatial–spectral FFPNet models. The spatial-only and spectral-only models correspond to the light-weight spatial FFP module and the spectral FFP module in Fig. 2 with fully connected classifiers, respectively.

Table 15 and Table 16, respectively, present the results of comparison of different models on the IP and UP datasets with sample patch sizes of d = 9, 15, 21, and 27 and 100 samples per category. Obviously, on both datasets, the performance of the spatial–spectral FFPNet model is significantly better than that of the exclusive spatial FFPNet and spectral FFPNet, especially when a small amount of spatial information is considered. Specifically, when d = 9, for the IP dataset, the OA value of the spatial–spectral model shows an improvement of 3.15% and 6.77% compared with the values achieved by the spatial-only and spectral-only models, respectively. In addition, the spatialonly and spectral-only models are more unstable and have limited accuracy for the UP dataset when spatial information is restricted (i.e., when d = 9, the spatial–spectral model shows an improvement of 12.39% and 33.26% compared with the values achieved by spatial-only and spectral-only models, respectively). Therefore, it is demonstrated that spatial information (the neighboring pixels) and spectral information should be simultaneously considered in the model to obtain an excellent classification result. Furthermore, the proposed spatial–spectral FFPNet model can effectively extract and fuse multiscale spatial and spectral features to achieve high classification accuracy even with a few training samples. However, Table 15 and Table 16 indicate that when the number of training samples is sufficient, the performance gap between the three models is not very large, especially for the IP dataset.

*Comparison with existing CNN methods.*

To further verify the effectiveness and superiority of the spatial–spectral FFPNet model, we compare it with some of the state-of-the-art and well-known CNN models developed in recent years for hyperspectral classification. The main comparisons about the configuration and training settings of these models are briefly described as follows:

Table 17: Classification accuracies (in %) of different CNN models developed from 2016 to 2019 and the proposed model for the IP dataset (with maximum epoch = 200 and mini-batch size = 24). The CNN by [38] is considered as the baseline.

| CNN Models | Attention Networks [43] | | Multiple CNN Fusion [40] | | | CNNs [39] | | | | | | CNN [38] | | | | spatial–spectral FFPNet | | | | |
|---|---|---|---|---|---|---|---|---|---|---|---|---|---|---|---|---|---|---|---|---|
| | A-ResNet | Samples | SODFN | FCLFN | Samples | 1-D | 2-D | 3-D | d = 9 | d = 19 | d = 29 | Samples | d = 9 | d = 19 | d = 29 | Samples | d = 9 | d = 15 | d = 19 | d = 29 | Samples |
| Alfalfa | 89.23 | 7 | 95.12 | 95.12 | 5 | 89.58 | 99.65 | 100.00 | 100.00 | 100.00 | 100.00 | 30 | 99.13 | 99.57 | 99.13 | 23 | 95.65 | 100.00 | 100.00 | 100.00 | 33 |
| Corn-notill | 97.69 | 214 | 98.91 | 99.38 | 143 | 85.68 | 90.64 | 96.34 | 90.57 | 94.06 | 97.17 | 150 | 80.48 | 94.47 | 98.17 | 200 | 98.78 | 99.35 | 99.59 | **100.00** | 200 |
| Corn-min | 99.29 | 125 | 99.06 | **100.00** | 83 | 87.36 | 99.11 | 99.49 | 97.69 | 96.43 | 98.17 | 150 | 96.65 | 98.22 | 98.92 | 200 | 99.84 | 98.25 | 99.05 | 100.00 | 200 |
| Corn | 92.24 | 36 | 98.12 | **100.00** | 24 | 93.33 | 100.00 | 100.00 | 99.92 | 100.00 | 100.00 | 100 | 99.66 | 100.00 | 100.00 | 118 | 100.00 | 100.00 | 100.00 | 100.00 | 181 |
| Grass/Pasture | 99.02 | 72 | 95.62 | 95.16 | 49 | 96.88 | 98.48 | 99.91 | 98.10 | 98.72 | 98.76 | 150 | 99.46 | 99.75 | 99.71 | 200 | 100.00 | 100.00 | 98.94 | 100.00 | 200 |
| Grass/Trees | 99.77 | 110 | 99.09 | 99.24 | 73 | 98.99 | 97.95 | 99.75 | 99.34 | 99.67 | 100.00 | 150 | 99.53 | 98.90 | 99.40 | 200 | 99.62 | **100.00** | 98.87 | 100.00 | 200 |
| Grass/pasture-mowed | 93.04 | 4 | 80.03 | 72.10 | 3 | 91.67 | 100.00 | 100.00 | 100.00 | 100.00 | 100.00 | 20 | 100.00 | 100.00 | 100.00 | 14 | 100.00 | 100.00 | 100.00 | 100.00 | 20 |
| Hay-windrowed | **100.00** | 72 | **100.00** | 99.53 | 48 | 99.49 | 100.00 | 100.00 | 99.58 | 99.92 | 100.00 | 150 | 99.67 | 99.62 | 100.00 | 200 | 100.00 | 100.00 | 100.00 | 100.00 | 200 |
| Oats | 90.59 | 3 | **100.00** | 88.89 | 2 | **100.00** | 100.00 | 100.00 | 100.00 | 100.00 | 100.00 | 15 | **100.00** | 100.00 | 100.00 | 14 | 100.00 | 100.00 | 100.00 | 100.00 | 14 |
| Soybeans-notill | 98.57 | 146 | 99.31 | 99.54 | 98 | 90.35 | 95.33 | 98.72 | 94.28 | 97.63 | 99.14 | 150 | 92.43 | 98.00 | 98.62 | 200 | 99.33 | 99.73 | 100.00 | 100.00 | 200 |
| Soybeans-min | 99.37 | 368 | 98.87 | 98.64 | 245 | 77.90 | 78.21 | 95.52 | 87.75 | 92.93 | 94.59 | 150 | 76.42 | 94.82 | 96.15 | 200 | 98.58 | 99.38 | **99.78** | 99.35 | 200 |
| Soybean-clean | 97.14 | 89 | 87.99 | 92.68 | 60 | 95.82 | 99.39 | 99.47 | 94.81 | 97.17 | 99.06 | 150 | 97.74 | 99.09 | 99.33 | 200 | 99.24 | 100.00 | 100.00 | 100.00 | 200 |
| Wheat | **100.00** | 31 | 97.83 | **100.00** | 21 | 98.59 | 100.00 | 100.00 | 100.00 | 100.00 | 100.00 | 150 | 99.71 | 100.00 | 99.90 | 102 | 100.00 | 100.00 | 100.00 | 100.00 | 143 |
| Woods | 99.57 | 190 | 99.91 | 99.91 | 126 | 98.55 | 97.71 | 99.55 | 98.09 | 97.88 | 99.76 | 150 | 97.71 | 99.85 | 98.96 | 200 | 98.78 | 99.34 | 100.00 | 100.00 | 200 |
| Bldg-Grass-Tree-Drives | 99.58 | 58 | 98.56 | 97.41 | 39 | 87.41 | 99.31 | 99.54 | 89.79 | 95.88 | 98.39 | 50 | 99.22 | 99.90 | 100.00 | 193 | 100.00 | 100.00 | 100.00 | 100.00 | 200 |
| Stone-steel | 97.72 | 14 | 96.39 | 97.59 | 10 | 98.06 | 99.22 | 99.34 | **100.00** | 99.57 | 98.92 | 50 | **100.00** | 100.00 | 100.00 | 46 | 95.74 | 100.00 | 100.00 | 100.00 | 75 |
| OA | 98.75 | | 98.21 | 98.56 | | 87.81 | 89.99 | 97.56 | 93.94 | 96.29 | 97.87 | | 90.11 | 97.23 | 98.37 | | 99.07 | 99.47 | 99.68 | **99.84** | |
| AA | 97.05 | | 96.54 | 95.94 | | 93.12 | 97.19 | 99.23 | 96.87 | 98.11 | 99.00 | | 96.12 | 98.79 | 99.27 | | 99.10 | 99.75 | 99.76 | **99.96** | |
| Kappa | 98.58 | | 97.97 | 98.36 | | 85.30 | 87.95 | 97.02 | 93.12 | 95.78 | 97.57 | | 88.81 | 96.85 | 98.15 | | 98.90 | 99.38 | 99.63 | **99.82** | |
| Total samples | | 1537 | | | 1029 | | | | | | | 1765 | | | | 2466 | | | | | 2306 |

Table 18: Classification accuracies (in %) of different CNN models developed from 2016 to 2019 and the proposed model for the UP dataset (with maximum epoch = 200, mini-batch size = 24). The CNN by [38] is considered the baseline.

| CNN Models | Attention Networks [43] | | Multiple CNN Fusion [40] | | | CNNs [39] | | | | | | CNN [38] | | | | Spatial-spectral FFPNet | | | | |
|---|---|---|---|---|---|---|---|---|---|---|---|---|---|---|---|---|---|---|---|---|
| | A-ResNet | Samples | SODFN | FCLFN | Samples | 1-D | 2-D | 3-D | d = 15 | d = 21 | d = 27 | Samples | d = 15 | d = 21 | d = 27 | Samples | d = 9 | d = 15 | d = 21 | d = 27 | Samples |
| Asphalt | 99.80 | 663 | 99.62 | 97.03 | 67 | 92.06 | 97.11 | 99.36 | 97.53 | 98.80 | 98.59 | 548 | 92.81 | 95.31 | 96.31 | 200 | 96.32 | 98.19 | 97.59 | **99.82** | 200 |
| Meadows | 99.97 | 1865 | 99.98 | **100.00** | 186 | 92.80 | 87.66 | 99.36 | 98.98 | 99.46 | 99.60 | 540 | 97.20 | 98.16 | 97.54 | 200 | 97.96 | 99.98 | 99.91 | 99.85 | 200 |
| Gravel | 99.56 | 210 | 97.47 | 95.14 | 22 | 83.67 | 99.69 | 99.69 | 98.96 | 99.59 | 99.45 | 392 | 96.97 | 97.92 | 96.84 | 200 | 95.99 | 98.66 | **99.81** | 99.05 | 200 |
| Trees | 99.74 | 306 | 91.79 | 88.49 | 31 | 93.85 | 98.49 | 99.63 | **99.75** | 99.68 | 99.57 | 542 | 98.62 | 98.74 | 97.58 | 200 | 99.60 | 96.61 | 96.34 | 96.61 | 200 |
| Painted metal sheets | 99.97 | 135 | 99.77 | 99.18 | 15 | 98.91 | **100.00** | 99.95 | 99.93 | 99.78 | 99.61 | 256 | 100.00 | 100.00 | 99.65 | 200 | 100.00 | 99.11 | 100.00 | 100.00 | 200 |
| Bare Soil | **100.00** | 503 | 96.77 | 99.46 | 51 | 94.17 | 98.00 | 99.96 | 99.42 | 99.93 | 99.84 | 532 | 98.57 | 99.57 | 99.33 | 200 | 93.95 | 100.00 | 100.00 | 100.00 | 200 |
| Bitumen | 99.16 | 133 | 89.51 | 95.89 | 15 | 92.68 | 99.89 | 100.00 | 99.89 | 98.71 | 99.88 | **100.00** | 375 | 97.27 | 99.75 | 98.90 | 200 | 100.00 | 100.00 | 100.00 | 100.00 | 200 |
| Self-Blocking Bricks | 99.73 | 368 | 97.59 | **100.00** | 38 | 89.09 | 99.70 | 99.65 | 98.58 | 99.53 | 99.67 | 514 | 96.17 | 98.20 | 98.89 | 200 | 96.30 | 99.02 | 98.04 | 99.13 | 200 |
| Shadows | 99.88 | 95 | 92.52 | 96.20 | 11 | 97.84 | 97.11 | 99.38 | 99.87 | 99.79 | 99.83 | 231 | 99.86 | 99.82 | 99.58 | 200 | 97.35 | 98.23 | **100.00** | 99.56 | 200 |
| OA | **99.86** | | 98.13 | 98.17 | | 92.28 | 94.04 | 99.54 | 98.87 | 99.47 | 99.48 | | 96.83 | 98.06 | 97.80 | | 96.51 | 99.25 | 99.15 | 99.53 | |
| AA | **99.76** | | 96.11 | 96.80 | | 92.55 | 97.52 | 99.66 | 99.08 | 99.60 | 99.57 | | 97.50 | 98.61 | 98.29 | | 96.23 | 98.87 | 99.08 | 99.33 | |
| Kappa | 99.82 | | 97.53 | 97.58 | | 90.37 | 92.43 | 99.41 | 98.51 | 99.30 | 99.32 | | 95.83 | 97.44 | 97.09 | | 95.37 | 99.01 | **98.87** | 99.38 | |
| Total samples | | 4278 | | | 436 | | | | | | | 3930 | | | | 1800 | | | | | 1800 |

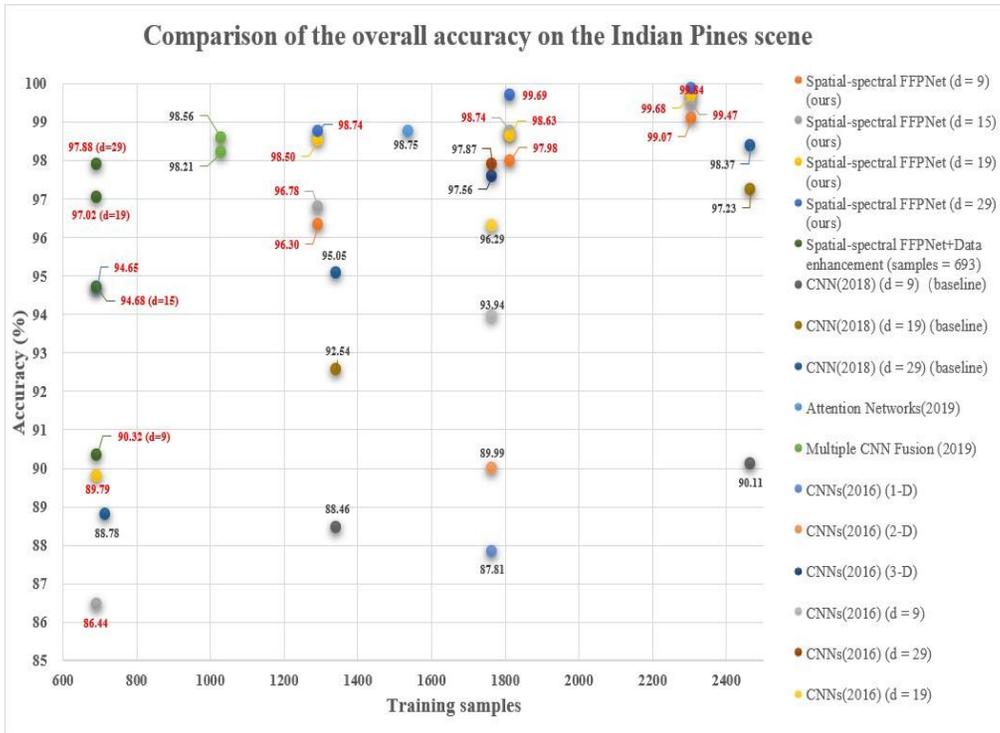

Figure 18: Comparison of the OA obtained by different CNN models for the IP dataset. The abscissa represents the total number of training samples (600–2500), and the ordinate represents the OA (%) of the CNN models. The figure mainly compares the accuracy of the existing CNN methods and the proposed spatial–spectral FFPNet under different training samples. The OA results obtained by our proposed model is presented in red, and those obtained by other CNNs [39], CNN [38], attention networks [43], and multiple CNN fusion [40] are presented in black.

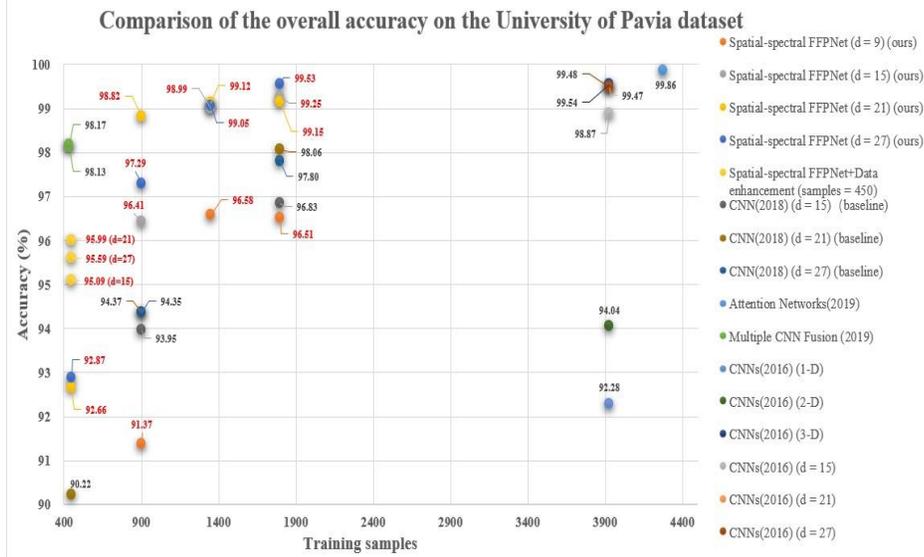

Figure 19: Comparison of the OA obtained by different CNN models for the UP dataset. The abscissa represents the total number of training samples (400–4400), and the ordinate represents the OA (%) of the CNN models. The figure mainly compares the accuracy of different models (existing CNN methods and the proposed spatial–spectral FFPNet) under different training samples. The OA results obtained by our proposed model are presented in red and those obtained by the existing CNNs [39], [38], attention networks [43], and multiple CNN fusion [40] are presented in black.

(1) CNNs [39]: First, the model configuration includes 1-D, 2-D, and 3-D CNNs. The 1-D CNN consists of five convolutional layers with ReLU and five pooling layers for the IP dataset as well as three convolutional layers with ReLU and three pooling layers for the UP dataset; the 1-D CNN extracts only spectral information. The 2-D CNN contains three 2D convolutional layers and two pooling layers. The latter two 2-D convolutional layers use the dropout strategy to prevent overfitting. The 3-D CNN is designed to effectively extract spatial and spectral information. It includes three 3D convolution and ReLU nonlinear activation layers, and the dropout strategy is also used to prevent overfitting. Overall, the design of the proposed model represents the early application of DCNN methods in hyperspectral image classification. However, although the models are simple and effective, *the spatial and spectral distribution diversities of hyperspectral datasets are not considered*. Second, in training settings, 1765 labeled samples were used as the training set for the IP dataset and 3930 samples as the training set for the UP dataset. Furthermore, experiments were conducted with different patch sizes (d = 9, 19, 29 for the IP dataset; d = 15, 21, 27 for the UP dataset) in the baseline [38].

(2) CNN by Paoletti et al. [38]: The CNN model serves as the baseline model for our hyperspectral experiments. First, in the model configuration, to extract hyperspectral classification features, three 3D convolutional layers (i.e., 600×5×5×200, 200×3×3×600, 200×1×1×200 for the IP dataset, and 380×7×7×103, 350×5×5×380, 350×1×1×350 for the UP dataset) are designed, and each convolution layer is followed by a ReLU function. To reduce the spatial resolution, the first two convolution layers are followed by two 2 × 2 max pooling. In addition, to prevent overfitting, the dropout method is executed in the first two convolution layers of the model, with probability = 0.1 for the IP dataset and 0.2 for the UP dataset. Next, a four-layer full connection classifies the extracted features. Although the 3D model requires less parameters and layers, *it cannot address the diversity problem of spatial object distribution in hyperspectral data and cannot make the full use of spectral information*. In addition, 3D convolution processes hyperspectral data are uniform volumetric data, while the hyperspectral actual object distribution is asymmetrical. Second, in the training setting, detailed experiments were conducted on different training samples and different patch sizes for the IP and UP datasets. The best experimental results of the baseline model (patch sizes d = 9, 19, and 29 with training samples = 2466 for the IP dataset; patch size d = 15, 21, and 27 with training samples = 1800 for the UP dataset) were considered for the comparison of the models.

(3) Attention networks [43]: A visual attention-driven mechanism applied to residual neural networks (ResNet) facilitates spatial–spectral hyperspectral image classification. Specifically, the attention mechanism is integrated into the residual part of ResNet, which mainly includes two parts, namely the trunk and mask. The trunk consists of some residual blocks that perform feature extraction from the data, while the mask consists of a symmetrical downsampler–upsampler structure to extract useful features from the current layer. Although the attention mechanism has been successfully applied to ResNet, this attention method *does not solve the problems of spatial distribution (the different geometric shapes of the objects) and spectral redundancy of hyperspectral data*. Second, the network was optimized using 1537 training samples with 300 epochs for the IP dataset and 4278 training samples with 300 epochs for the UP dataset.

(4) Multiple CNN fusion [40]: Compared with other models, although the multiscale spectral and spatial feature fusion model is time consuming, it can achieve superior classification accuracy and hence has been gaining prominence in hyperspectral image classification. For example, Zhao et al. [40] presented a multiple convolutional layers fusion framework, which fuses features extracted from different convolutional layers for hyperspectral image classification. *This multiple CNN model only considers the fusion of spatial features at different scales, but not the effective extraction of spatial and spectral features at multiple scales.* Specifically, the multiscale spectral and spatial feature fusion model is divided into two types according to the fusion mechanism. The first one is the side output decision fusion network (SODFN), which applies majority voting to many side classification maps generated by each convolutional layer. The other one is the fully convolutional layer fusion network (FCLFN), which combines all features generated by each convolutional layer. Second, the SODFN and FCLFN parameters were tuned using 1029 training samples for the IP dataset and 436 training samples for the UP dataset.

**Indian Pines dataset benchmark evaluation:** the classification results for the IP dataset obtained by different CNN models and our proposed model are shown in Table 17. Obviously, the spatial–spectral FFPNet without data enhancement generates the highest OA, AA, kappa coefficient and thus presents the best performance among all benchmark models. The proposed model also shows excellent performance in each class. The best result of the spatial–spectral FFPNet exceeds the best result of the baseline model [38] by 1.47% in terms of OA. Notably, the spatial–spectral FFPNet shows superior performance in all experimental configurations compared with the baseline model. In particular, when spatial information is limited (i.e., patch size d = 9), the proposed model outperforms the same configuration of the baseline model by 8.96%, and even exceeds the best configuration of the baseline model (d = 29, training samples = 2466) by 0.7%. The results presented in Table 17 further demonstrates the superiority of the proposed spatial– spectral FFPNet and its robustness in case of a small number of training samples.

A graphical comparison of the OA values for the IP dataset obtained by different CNN models is shown in Fig. 18. The OA results obtained by our proposed model are presented in red, while those obtained by the benchmark models are presented in black. Clearly, our model shows more promising performance under different training samples compared with the well-known hyperspectral classification CNN models. Specifically, when the number of training samples is relatively small (600–800), the spatial–spectral FFPNet with data enhancement acquires outstanding performance. As the number of samples increases, the dominance of the spatial–spectral FFPNet over the other methods is relatively more. In addition, note that attention networks [43] and multiple CNN fusion [40] perform better than the CNN models by [39] and [38]; this is also in line with the current development trend of hyperspectral classification; that is, the application of multiscale feature fusion and attention mechanisms in the spatial and spectral dimensions.

**University of Pavia dataset benchmark evaluation:** Table 18 lists the classification results of different CNN models developed from 2016 to 2019 and our proposed model for the UP dataset. The spatial–spectral FFPNet (without data enhancement) with different d values shows outstanding results in performance. Specifically, with the experimental configuration of d = 27, the proposed model shows an improvement of 1.73% compared to the CNN model [38] in terms of OA. The spatial–spectral FFPNet also shows a great performance compared with the baseline [38] with the same experimental configuration. Furthermore, the performance of the spatial–spectral FFPNet with a lower patch size (d = 9) differs slightly from that of the baseline model [38] with d = 15. Notably, attention networks [43] perform the best among all model in terms of OA and AA results because of the sufficient training samples (4278), while our proposed model performs the best in terms of the kappa coefficient.

Fig. 19 provides a graphical comparison of the OA of different CNN models for the UP dataset. Again, our model attains more homogenized and favorable classification results. As Fig. 19 shows, the results of our proposed model are centered around 400–1900 training samples. However, the local OA comparison chart indicates that the multiple CNN fusion [40] is superior (98.17%) even for an extremely small number of training samples (approximately 400). As the training samples increase, the superiority of the spatial–spectral FFPNet gradually gains prominence. In addition, for a large training sample (>3000), attention networks [43] perform considerably better than the traditional CNN model [39].

## 4. Conclusions

In this study, we mainly focus on spatial object distribution diversity and spectral information extraction, which are the major challenges of highresolution and hyperspectral remote sensing images. To address the spatial and spectral problems, three novel and practical attention-based modules were proposed: attention-based multiscale fusion, region pyramid attention, and adaptive-ASPP. We constructed different forms of feature fusion pyramid frameworks (two-layer or three-layer pyramids) by combining these attentionbased modules. First, we developed a new semantic segmentation framework for high-resolution images, called the heavy-weight spatial FFPNet. Second, for the classification of hyperspectral images, an end-to-end spatial–spectral FFPNet was presented to extract and fuse multiscale spatial and spectral features. The experiments conducted on two high-resolution datasets demonstrated that the proposed heavy-weight spatial FFPNet achieves excellent segmentation accuracy. Detailed ablation studies further revealed the superiority of the three attention-based modules in processing the spatial distribution diversity of remote sensing images. Furthermore, detailed training parameter analysis and comparison with other state-of-the-art CNNs (such as [38]) were performed on the two hyperspectral datasets. The results demonstrated that the spatial–spectral FFPNet is more robust and achieves greater accuracy in case when the number of training samples of the hyperspectral dataset is small and that it can obtain state-of-the-art results under different training samples. Overall, the proposed methods can serve as a new baseline for remote sensing image segmentation and classification. In future work, we will focus on few-shot or zero-shot segmentation and classification in high-resolution or hyperspectral remote sensing data to promote practical application of deep learning in remote sensing image perception.

**References**

[1] P. Ghamisi, M. D. Mura, and J. A. Benediktsson, A survey on spectralspatial classification techniques based on attribute profiles, IEEE Transactions on Geoscience and Remote Sensing, vol. 53, no. 5, pp. 2335C2353, 2015.

[2] N. Wang, F. Chen, B. Yu, and Y. Qin, Segmentation of large-scale remotely sensed images on a spark platform: A strategy for handling massive image tiles with the mapreduce model, Isprs Journal of Photogrammetry and Remote Sensing, vol. 162, pp. 137C147, 2020.

[3] W. Sun and R. Wang, Fully convolutional networks for semantic segmentation of very high resolution remotely sensed images combined with dsm, IEEE Geoscience and Remote Sensing Letters, vol. 15, no. 3, pp. 474C478, 2018.

[4] Y. Liu, B. Fan, L. Wang, J. Bai, S. Xiang, and C. Pan, Semantic labeling in very high resolution images via a self-cascaded convolutional neural network, Isprs Journal of Photogrammetry and Remote Sensing, vol. 145, pp. 78C95, 2017.

[5] N. Audebert, B. Le Saux, and S. Lef`evre, Deep learning for classification of hyperspectral data: A comparative review, IEEE Geoscience and Remote Sensing Magazine, vol. 7, no. 2, pp. 159C173, 2019.

[6] F. Melgani and L. Bruzzone, Classification of hyperspectral remote sensing images with support vector machines, IEEE Transactions on Geoscience and Remote Sensing, vol. 42, no. 8, pp. 1778C1790, 2014.

[7] T. Lin, P. Dollar, R. Girshick, K. He, B. Hariharan, and S. Belongie, Feature pyramid networks for object detection, pp. 936C944, 2017.

[8] J. Long, E. Shelhamer, and T. Darrell, Fully convolutional networks for semantic segmentation, in Proceedings of the IEEE conference on computer vision and pattern recognition, 2015, pp. 3431C3440.

[9] H. Wang, Y. Wang, Q. Zhang, S. Xiang, and C. Pan, Gated convolutional neural network for semantic segmentation in high-resolution images, Remote Sensing, vol. 9, no. 5, p. 446, 2017.

[10] G. Cheng, Y. Wang, S. Xu, H. Wang, S. Xiang, and C. Pan, Automatic road detection and centerline extraction via cascaded end-to-end convolutional neural network, IEEE Transactions on Geoscience and Remote Sensing, vol. 55, no. 6, pp. 3322C3337, 2017.

[11] Y. Liu, B. Fan, L. Wang, J. Bai, S. Xiang, and C. Pan, Semantic labeling in very high resolution images via a self-cascaded convolutional neural network, Isprs Journal of Photogrammetry and Remote Sensing, vol.
145, pp. 78C95, 2018.

[12] W. Cheng, W. Yang, M. Wang, G. Wang, and J. Chen, Context aggregation network for semantic labeling in aerial images, Remote Sensing, vol. 11, no. 10, p. 1158, 2019.

[13] L. Mou, Y. Hua, and X. X. Zhu, A relation-augmented fully convolutional network for semantic segmentation in aerial scenes, in Proceedings of the

IEEE conference on computer vision and pattern recognition, 2019, pp. 12 416C12 425.

[14] P. Li, Y. Lin, and E. Schultz-Fellenz, Contextual hourglass network for semantic segmentation of high resolution aerial imagery, arXiv preprint arXiv:1810.12813, 2018.

[15] X. Yang, J. Yang, J. Yan, Y. Zhang, T. Zhang, Z. Guo, X. Sun, and K. Fu, Scrdet: Towards more robust detection for small, cluttered and rotated objects, in Proceedings of the IEEE International Conference on Computer Vision, 2019, pp. 8232C8241.

[16] C. Sebastian, R. Imbriaco, E. Bondarev, and P. H. de With, Adversarial loss for semantic segmentation of aerial imagery, arXiv preprint arXiv:2001.04269, 2020.

[17] R. Dong, X. Pan, and F. Li, Denseu-net-based semantic segmentation of small objects in urban remote sensing images, IEEE Access, vol. 7, pp. 65 347C65 356, 2019.

[18] Y. Du, W. Song, Q. He, D. Huang, A. Liotta, and C. Su, Deep learning with multi-scale feature fusion in remote sensing for automatic oceanic eddy detection, Information Fusion, vol. 49, pp. 89C99, 2019.

[19] S. Jain and B. C. Wallace, Attention is not explanation, arXiv preprint arXiv:1902.10186, 2019.

[20] X. Wang, R. Girshick, A. Gupta, and K. He, Non-local neural networks, in Proceedings of the IEEE conference on computer vision and pattern recognition, 2018, pp. 7794C7803.

[21] J. Fu, J. Liu, H. Tian, Y. Li, Y. Bao, Z. Fang, and H. Lu, Dual attention network for scene segmentation, in Proceedings of the IEEE Conference on Computer Vision and Pattern Recognition, 2019, pp. 3146C3154.

[22] Z. Zhu, M. Xu, S. Bai, T. Huang, and X. Bai, Asymmetric non-local neural networks for semantic segmentation, in Proceedings of the IEEE International Conference on Computer Vision, 2019, pp. 593C602.

[23] O. Oktay, J. Schlemper, L. L. Folgoc, M. Lee, M. Heinrich, K. Misawa, K. Mori, S. McDonagh, N. Y. Hammerla, B. Kainz et al., Attention u-net:

Learning where to look for the pancreas, arXiv preprint arXiv:1804.03999, 2018.

[24] Z. Huang, X. Wang, L. Huang, C. Huang, Y. Wei, and W. Liu, Ccnet: Criss-cross attention for semantic segmentation, in Proceedings of the IEEE International Conference on Computer Vision, 2019, pp. 603C612.

[25] F. Zhang, Y. Chen, Z. Li, Z. Hong, J. Liu, F. Ma, J. Han, and E. Ding, Acfnet: Attentional class feature network for semantic segmentation, in Proceedings of the IEEE International Conference on Computer Vision, 2019, pp. 6798C6807.

[26] V. A. Sindagi and V. M. Patel, Multi-level bottom-top and top-bottom feature fusion for crowd counting, in Proceedings of the IEEE International Conference on Computer Vision, 2019, pp. 1002C1012.

[27] R. Niu, HMANet: Hybrid multiple attention network for semantic segmentation in aerial images, arXiv preprint arXiv:2001.02870, 2020.

[28] C. Guo, B. Fan, Q. Zhang, S. Xiang, and C. Pan, Augfpn: Improving multi-scale feature learning for object detection, arXiv preprint arXiv:1912.05384, 2019.

[29] X. Jin, C. Lan, W. Zeng, Z. Zhang, and Z. Chen, Casenet: contentadaptive scale interaction networks for scene parsing, arXiv preprint arXiv:1904.08170, 2019.

[30] Y. Tarabalka, J. A. Benediktsson, and J. Chanussot, Spectralspatial classification of hyperspectral imagery based on partitional clustering techniques, IEEE Transactions on Geoscience and Remote Sensing, vol. 47, no. 8, pp. 2973C2987, 2009.

[31] R. Archibald and G. Fann, Feature selection and classification of hyperspectral images with support vector machines, IEEE Geoscience and Remote Sensing Letters, vol. 4, no. 4, pp. 674C677, 2007.

[32] S. Sun, P. Zhong, H. Xiao, and R. Wang, Active learning with gaussian process classifier for hyperspectral image classification, IEEE Transactions on Geoscience and Remote Sensing, vol. 53, no. 4, pp. 1746C1760, 2015.

[33] Y. Chen, N. M. Nasrabadi, and T. D. Tran, Hyperspectral image classification using dictionary-based sparse representation, IEEE Transactions on Geoscience and Remote Sensing, vol. 49, no. 10, pp. 3973C3985, 2011.

[34] K. Makantasis, K. Karantzalos, A. Doulamis, and N. Doulamis, Deep supervised learning for hyperspectral data classification through convolutional neural networks, pp. 4959C4962, 2015.

[35] W. Zhao and S. Du, Spectralspatial feature extraction for hyperspectral image classification: A dimension reduction and deep learning approach, IEEE Transactions on Geoscience and Remote Sensing, vol. 54, no. 8, pp. 4544C4554, 2016.

[36] Y. Luo, J. Zou, C. Yao, T. Li, and G. Bai, Hsi-cnn: A novel convolution neural network for hyperspectral image. arXiv: Computer Vision and Pattern Recognition, 2018.

[37] Y. Li, H. Zhang, and Q. Shen, Spectralspatial classification of hyperspectral imagery with 3d convolutional neural network, Remote Sensing, vol. 9, no. 1, p. 67, 2017.

[38] M. Paoletti, J. Haut, J. Plaza, and A. Plaza, A new deep convolutional neural network for fast hyperspectral image classification, ISPRS journal of photogrammetry and remote sensing, vol. 145, pp. 120C147, 2018.

[39] Y. Chen, H. Jiang, C. Li, X. Jia, and P. Ghamisi, Deep feature extraction and classification of hyperspectral images based on convolutional neural networks, IEEE Transactions on Geoscience and Remote Sensing, vol. 54, no. 10, pp. 6232C6251, 2016.

[40] G. Zhao, G. Liu, L. Fang, B. Tu, and P. Ghamisi, Multiple convolutional layers fusion framework for hyperspectral image classification, Neurocomputing, vol. 339, pp. 149C160, 2019.

[41] Z. Gong, P. Zhong, Y. Yu, W. Hu, and S. Li, A cnn with multiscale convolution and diversified metric for hyperspectral image classification, IEEE Transactions on Geoscience and Remote Sensing, vol. 57, no. 6, pp. 3599C3618, 2019.

[42] M. Imani and H. Ghassemian, An overview on spectral and spatial information fusion for hyperspectral image classification: Current trends and challenges, Information Fusion, vol. 59, pp. 59C83, 2020.

[43] J. M. Haut, M. E. Paoletti, J. Plaza, A. Plaza, and J. Li, Visual attention-driven hyperspectral image classification, IEEE Transactions on Geoscience and Remote Sensing, vol. 57, no. 10, pp. 8065C8080, 2019.

[44] X. Mei, E. Pan, Y. Ma, X. Dai, J. Huang, F. Fan, Q. Du, H. Zheng, and J. Ma, Spectral-spatial attention networks for hyperspectral image classification, Remote Sensing, vol. 11, no. 8, p. 963, 2019.

[45] Q. Xu, C. Ouyang, T. Jiang, X. Fan, and D. Cheng, DFPENet-geology: A deep learning framework for high precision recognition and segmentation of co-seismic landslides, arXiv preprint arXiv:1908.10907, 2019.

[46] L.-C. Chen, Y. Zhu, G. Papandreou, F. Schroff, and H. Adam, Encoderdecoder with atrous separable convolution for semantic image segmentation, in Proceedings of the European conference on computer vision (ECCV), 2018, pp. 801C818.

[47] J. Deng, W. Dong, R. Socher, L.-J. Li, K. Li, and L. Fei-Fei, Imagenet: A large-scale hierarchical image database, in 2009 IEEE conference on computer vision and pattern recognition. IEEE, 2009, pp. 248C255.

[48] O. Ronneberger, P. Fischer, and T. Brox, U-net: Convolutional networks for biomedical image segmentation, in International Conference on Medical image computing and computer-assisted intervention. Springer, 2015, pp. 234C241.

[49] D. Lin, D. Shen, S. Shen, Y. Ji, D. Lischinski, D. Cohen-Or, and H. Huang, Zigzagnet: Fusing top-down and bottom-up context for object segmentation, in Proceedings of the IEEE Conference on Computer Vision and Pattern Recognition, 2019, pp. 7490C7499.

[50] M. Zhen, J. Wang, L. Zhou, T. Fang, and L. Quan, Learning fully dense neural networks for image semantic segmentation, in Proceedings of the AAAI Conference on Artificial Intelligence, vol. 33, 2019, pp. 9283C 9290.

[51] Z. Zhang, X. Zhang, C. Peng, X. Xue, and J. Sun, Exfuse: Enhancing feature fusion for semantic segmentation, in Proceedings of the European Conference on Computer Vision (ECCV), 2018, pp. 269C284.

[52] X. Li, H. Zhao, L. Han, Y. Tong, and K. Yang, Gff: Gated fully fusion for semantic segmentation, arXiv preprint arXiv:1904.01803, 2019.

[53] Z. Zhou, M. M. R. Siddiquee, N. Tajbakhsh, and J. Liang, Unet++: Redesigning skip connections to exploit multiscale features in image segmentation, IEEE Transactions on Medical Imaging, 2019.

[54] H. Zhao, Y. Zhang, S. Liu, J. Shi, C. Change Loy, D. Lin, and J. Jia, Psanet: Point-wise spatial attention network for scene parsing, in Proceedings of the European Conference on Computer Vision (ECCV), 2018, pp. 267C283.

[55] L.-C. Chen, G. Papandreou, I. Kokkinos, K. Murphy, and A. L. Yuille, Deeplab: Semantic image segmentation with deep convolutional nets, atrous convolution, and fully connected crfs, IEEE transactions on pattern analysis and machine intelligence, vol. 40, no. 4, pp. 834C848, 2017.

[56] K. He, X. Zhang, S. Ren, and J. Sun, Deep residual learning for image recognition, in Proceedings of the IEEE conference on computer vision and pattern recognition, 2016, pp. 770C778.

[57] K. Simonyan and A. Zisserman, Very deep convolutional networks for large-scale image recognition, 2014.

[58] M. Yang, K. Yu, C. Zhang, Z. Li, and K. Yang, Denseaspp for semantic segmentation in street scenes, in Proceedings of the IEEE Conference on Computer Vision and Pattern Recognition, 2018, pp. 3684C3692.

[59] M. Volpi and D. Tuia, Dense semantic labeling of subdecimeter resolution images with convolutional neural networks, IEEE Transactions on Geoscience and Remote Sensing, vol. 55, no. 2, pp. 881C893, 2016.

[60] N. Audebert, B. Le Saux, and S. Lef'evre, Beyond rgb: Very high resolution urban remote sensing with multimodal deep networks, ISPRS Journal of Photogrammetry and Remote Sensing, vol. 140, pp. 20C32, 2018.

[61] D. Marmanis, K. Schindler, J. D. Wegner, S. Galliani, M. Datcu, and U. Stilla, Classification with an edge: Improving semantic image segmentation with

boundary detection, ISPRS Journal of Photogrammetry and Remote Sensing, vol. 135, pp. 158C172, 2018.

[62] H. Zhao, J. Shi, X. Qi, X. Wang, and J. Jia, Pyramid scene parsing network, in Proceedings of the IEEE conference on computer vision and pattern recognition, 2017, pp. 2881C2890.

[63] V. Badrinarayanan, A. Kendall, and R. Cipolla, Segnet: A deep convolutional encoder-decoder architecture for image segmentation, IEEE transactions on pattern analysis and machine intelligence, vol. 39, no. 12, pp. 2481C2495, 2017.

[64] J. Sherrah, Fully convolutional networks for dense semantic labelling of high-resolution aerial imagery, arXiv preprint arXiv:1606.02585, 2016.

[65] B. Pan, Z. Shi, X. Xu, T. Shi, N. Zhang, and X. Zhu, Coinnet: Copy initialization network for multispectral imagery semantic segmentation, IEEE Geoscience and Remote Sensing Letters, vol. 16, no. 5, pp. 816C820, 2018.

[66] D. P. Kingma and J. Ba, Adam: A method for stochastic optimization, arXiv preprint arXiv:1412.6980, 2014.